\documentclass{l4dc2027}

\title[Physics-Grounded STL Monitoring from Image Observations]{Physics-Grounded STL Monitoring from Image Observations}
\usepackage{times}
\usepackage[rightcaption]{sidecap}
\usepackage{xcolor}

\newcommand*{\hist}{\square^{-}}%
\newcommand*{\once}{\lozenge^{-}}%
\newcommand*{\InferenceCost}{41~ms with bf16 arithmetic, below the 50~ms frame period, and batched offline decoding runs at 55 frames per second}

\makeatletter
\g@addto@macro\normalsize{%
  \setlength{\abovedisplayskip}{5pt plus 2pt minus 2pt}%
  \setlength{\belowdisplayskip}{5pt plus 2pt minus 2pt}%
  \setlength{\abovedisplayshortskip}{2pt plus 1pt}%
  \setlength{\belowdisplayshortskip}{3pt plus 1pt minus 1pt}}
\renewcommand{\section}{\@startsection{section}{1}{\z@}%
   {-0.16in \@plus -0.5ex \@minus -.2ex}{0.06in \@plus.2ex}{\normalfont\rmfamily\bfseries\large\raggedright}}
\renewcommand\subsection{\@startsection{subsection}{2}{\z@}%
   {-0.13in \@plus -0.5ex \@minus -.2ex}{0.05in \@plus .2ex}{\normalfont\rmfamily\bfseries\normalsize\raggedright}}
\renewcommand\paragraph{\@startsection{paragraph}{4}{\z@}%
   {0.8ex plus 0.3ex minus .2ex}{-1em}{\normalfont\normalsize\rmfamily\bfseries}}
\makeatother

\makeatletter
\@ifundefined{problem}{\newtheorem{problem}{Problem}}{}
\@ifundefined{remark}{}{}
\@ifundefined{theorem}{}{}
\@ifundefined{lemma}{}{}
\@ifundefined{definition}{\newtheorem{definition}{Definition}}{}
\@ifundefined{corollary}{}{}
\@ifundefined{proposition}{\newtheorem{proposition}{Proposition}}{}
\makeatother

\coltauthor{%
 \Name{Xinpei Ni} \Email{xni32@gatech.edu}\\
 \Name{Glen Chou} \Email{chou@gatech.edu}\\
 \Name{Samuel Coogan} \Email{sam.coogan@gatech.edu}\\
 \addr Georgia Institute of Technology%
}

\begin{document}

\maketitle
\vspace{-3.5em}
\begin{abstract}%
Robotic manipulation missions must satisfy timed and conditional requirements that can be expressed in signal temporal logic (STL). While a practical means of monitoring these requirements is from camera images, this is challenging because STL robustness depends on physical margins, such as distances and clearances, that are not directly observed. Rather than learning a separate visual monitor for each formula, we learn a shared latent representation from multi-view images and decode physically meaningful margins for individual predicates.
Co-training the encoder with predicate decoders increases the certification rate from 58.7\% to 96.3\% compared with a frozen encoder, despite nearly unchanged predicate sign accuracy.
STL semantics then compose these margins into robustness estimates for any bounded formula over the trained predicates, without retraining. We further calibrate the decoding errors of selected STL clauses using conformal prediction, yielding two-sided intervals that certify both satisfaction and violation and propagate through interval STL to any formula composed from the clauses.
In simulated kitchen missions (LIBERO), our monitor correctly predicts satisfaction or violation in 98.9\% of 4662 evaluations across 18 unseen formulas. For missions lasting up to 17 min, calibrated bounds establish satisfaction or violation for 96.9\% of missions without incorrect conclusions. A single calibration also provides simultaneous 91.0\% coverage across all 255 conjunctions of eight clauses, allowing new formulas without recalibration.
\end{abstract}

\vspace{-1em}
\section{Introduction}\label{sec:intro}
Robotic manipulation tasks often impose temporal or conditional requirements in addition to a terminal outcome.
Kitchen tasks, for example, bound how long a cooking step may last, so that the food is neither under- nor over-cooked.
Signal temporal logic (STL) is a formal language for such requirements~\citep{maler04stl}, and its quantitative semantics assigns each trajectory a robustness value, whose sign indicates satisfaction or violation and whose magnitude is the margin~\citep{donze10, fainekos09}.
In this work, we monitor STL specifications over minutes-long manipulation missions that are observed only through camera images.
This setting is demanding because the robustness depends on physical quantities, such as distances and clearances, that are not measured directly and must be inferred from the images.

Concretely, a monitor for this setting faces three challenges.
First, robustness must be estimated from images, whereas classical STL monitors compute it from perfect state information~\citep{donze10, fainekos09, deshmukh17robust}, which is rarely available on a robot.
Second, the monitor should be able to support novel unseen formulas, since the requirements of a mission change more often than the perception model can be feasibly retrained.
Third, the monitor should provide calibrated bounds containing the true robustness of these formulas, enabling reliable conclusions about satisfaction or violation.
Existing camera-based monitors face three key limitations.
First, \emph{one-sided certificates}: \citet{hoxha26} certify short-horizon, negation-free formulas from below only, so violations are never certified.
Second, \emph{lack of physically meaningful robustness}: other visual monitors~\citep{etl26,nsvstl} measure similarity to satisfying examples rather than physical margins to constraint violation. Consequently, their robustness estimates do not directly indicate how close the robot is to violating a physical requirement. Third, \emph{data-inefficient calibration}: conformal prediction (CP)~\citep{angelopoulos23} provides probabilistic guarantees, but two-sided calibration for a formula does not extend to other formulas. Certified monitoring of many specifications therefore requires separate calibration or conservative union bounds, making guarantees impractical when calibration data is limited.

To address these limitations, we propose an image-based STL monitor that estimates physically meaningful robustness, supports new specifications without retraining, and provides reusable probabilistic guarantees. First, we define each predicate through a signed physical margin computed from the simulator state and train a decoder for each predicate on a shared latent representation of multi-view images, which reduces occlusions. By learning to estimate individual predicate margins rather than the robustness of a particular formula, we can then use STL semantics to compositionally estimate the robustness of any bounded formula over the trained predicates, without retraining. Finally, to avoid recalibrating the monitor for every new formula, our key insight is to calibrate the decoding errors of a fixed set of clauses (STL subformulas) over entire missions, rather than calibrating each formula separately. We propagate the resulting clause-level intervals through STL operators using interval STL (I-STL)~\citep{baird2023istl}. Assuming the calibration and test missions are exchangeable, we show that a single calibration simultaneously bounds the true robustness of every formula composed from these clauses with probability at least $1-\alpha$.

\noindent\textbf{Related work.}
Classical online STL monitors bound the robustness of partial trajectories~\citep{deshmukh17robust}, but assume direct access to physical state measurements, limiting their applicability to vision-based robotic systems.
Perception-oriented temporal logics~\citep{balakrishnan19tqtl, hekmatnejad24formalizing} specify requirements on perception outputs, while vision-language-based monitors~\citep{liu23reflect, duan25aha, zhou25codeasmonitor} detect manipulation failures. However, these approaches do not provide physically meaningful STL robustness estimates with calibrated guarantees.
Interval STL~\citep{zhong21extending, baird2023istl} propagates uncertainty through temporal logic operators, but assumes that bounds on the underlying signals are already available.
More closely related are
\citet{hoxha26}, whose single calibration certifies every formula in a negation-free fragment, but only from below. Formula-level monitors \citep{lindemann23,cairoli23} are two-sided
but formula-specific. Our clause-level calibration is both two-sided and reusable.

In contrast, we learn physically grounded predicate margins from images and calibrate the decoding errors of a fixed set of clauses rather than individual specifications. By composing these margins using STL semantics and propagating clause-level intervals using I-STL, our approach provides two-sided robustness bounds for any formula composed from the calibrated clauses, without retraining or recalibration. This enables reusable probabilistic guarantees for long-horizon missions observed only through cameras.

\noindent\textbf{Contributions.}
We make four contributions:
\vspace{-.5em}
\begin{itemize}
\item \textit{Physically grounded predicate decoding:} We develop a multi-view latent encoder with per-predicate decoders that estimate signed physical margins from images (Sec.~\ref{subsec:pipeline}).
\vspace{-.5em}
\item \textit{Composable STL monitoring:} By learning predicate margins independently of the specification, our monitor estimates the robustness of any bounded STL formula over the trained predicates, without retraining, for both offline and online monitoring (Sec.~\ref{sec:problem} and~\ref{subsec:pipeline}).
\vspace{-.5em}
\item \textit{Reusable probabilistic guarantees:} We prove that one clause-level calibration, propagated through I-STL, simultaneously bounds the robustness of every formula over the calibrated clauses, including online at every monitored time step. We also identify the image frame determining each robustness estimate to facilitate diagnosis (Sec.~\ref{subsec:calibration}).
\vspace{-.5em}
\item \textit{Experimental validation:} On LIBERO missions~\citep{libero} with unseen specifications, new scenes, and durations up to 17~min, we outperform visual STL monitors, embedding-based predicates, and vision-language models in monitoring and certification (Sec.~\ref{sec:experiments_setup} and~\ref{sec:experiments_results}). We also show that encoder-decoder co-training can dramatically improve the usefulness of statistical certificates without dramatically changing sign accuracy.
\end{itemize}
The remainder of this paper states the monitoring problem in Section~\ref{sec:problem}, presents the method in Section~\ref{sec:method}, describes the experiments in Sections~\ref{sec:experiments_setup} and~\ref{sec:experiments_results}, followed by a conclusion in Section~\ref{sec:conclusion}.
\vspace{-0.5em}

\section{Problem Formulation}\label{sec:problem}
\vspace{-0.25em}
\paragraph{Missions and predicates.}
A mission is a pair $(x_{0:T},y_{0:T})$, where $x_t\in\mathcal X$ is the true physical state and $y_t\in\mathcal Y$ is the image recorded at time step $t\in\{0,\ldots,T\}$.
Each of the $N$ predicates $\pi_i$, $i\in\{1,\ldots,N\}$, holds at $t$ if and only if its predicate function $\mu_i:\mathcal X\to[-1,1]$, a signed physical margin divided by a tolerance and clipped (Appendix~\ref{app:moka}), satisfies $\mu_i(x_t)\ge0$.
Since the monitor sees only images, we estimate each $\mu_i$ from the images (Section~\ref{subsec:pipeline}).

\paragraph{STL.}
We specify requirements in past-time STL~\citep{maler04stl,maler05past},
\begin{equation}\label{eq:syntax}
  \varphi::=\pi_i\mid\neg\varphi\mid\varphi_1\wedge\varphi_2\mid\varphi_1\vee\varphi_2\mid\hist_I\varphi\mid\once_I\varphi\mid\varphi_1\mathcal S_I\varphi_2,
\end{equation}
where $\hist$, $\once$, and $\mathcal S$ are the operators historically, once, and since, and each interval $I=[a,b]\cap\mathbb Z_{\ge0}$, $0\le a\le b\le\infty$, specifies how far back the operator looks.
The robustness $\rho(\varphi,t)$ of $\varphi$ at $t$ is defined recursively~\citep{dokhanchi14online}, with $t-I=\{t-\tau:\tau\in I\}$ and $k\in\{1,2\}$, by
\begingroup\small
\begin{equation}\label{eq:semantics}
  \begin{aligned}
    &\rho(\pi_i,t)=\mu_i(x_t),\;\;\rho(\neg\varphi,t)=-\rho(\varphi,t),\;\;
    \rho(\varphi_1\wedge\varphi_2,t)=\min_k\rho(\varphi_k,t),\;\;\rho(\varphi_1\vee\varphi_2,t)=\max_k\rho(\varphi_k,t),\\
    &\textstyle\rho(\hist_I\varphi,t)=\min_{t'\in t-I}\rho(\varphi,t'),\;\;\rho(\once_I\varphi,t)=\max_{t'\in t-I}\rho(\varphi,t'),\\
    &\textstyle\rho(\varphi_1\mathcal S_I\varphi_2,t)=\max_{t'\in t-I}\min\bigl\{\rho(\varphi_2,t'),\min_{t'<u\le t}\rho(\varphi_1,u)\bigr\}.
  \end{aligned}\hspace{-15pt}
\end{equation}
\endgroup
\looseness-1The sign of $\rho(\varphi,t)$ indicates if the formula $\varphi$ is satisfied at time $t$: positive values indicate satisfaction and negative values indicate violation.
A bounded window is evaluated only when it lies within the recorded steps, and an unbounded window ($b=\infty$) reaches back to step 0; a window may depend on the mission length $T$, e.g., $\once_{[T-a,T]}$ evaluated at $T$ reads the first $a$ steps.
Let $\|\varphi\|$ denote the horizon of $\varphi$, i.e., the number of past time steps required to evaluate it~\citep{dokhanchi14online}.
Replacing the true predicate margins $\rho(\pi_i,t)=\mu_i(x_t)$ in~\eqref{eq:semantics} with estimates $\hat\rho(\pi_i,t)$ obtained from the image $y_t$ yields an estimate $\hat\rho(\varphi,t)$ of the formula robustness, computed from the images up to $t$ without requiring perfect state information. Since the robot observes only camera images rather than the true physical state, our goal is to estimate STL robustness and provide probabilistic guarantees that enable reliable conclusions about satisfaction or violation (Prob. \ref{prob:monitoring}). To provide these guarantees, we seek to calibrate robustness estimation error, either for a fixed STL formula or for a set of clauses whose bounds can be reused across multiple formulas (Prob. \ref{prob:calibration}).
\vspace{-.25em}
\begin{problem}[Probabilistically-certified monitoring]\label{prob:monitoring}
Given an STL formula $\varphi$ and the images $y_{0:t}$ recorded up to a query time step $t$, design a monitor that returns
(i) a point estimate $\hat\rho_t$ of the robustness $\rho_t$ that the query asks about;
(ii) a robustness interval $\mathcal I_t$, computed only from $y_{0:t}$ and a calibration set, such that, for a miscoverage level $\alpha\in(0,1)$, with probability at least $1-\alpha$ the robustness $\rho_t$ lies in $\mathcal I_t$ for every queried $t$; and
(iii) a conclusion, \emph{satisfied} if $\inf\mathcal I_t>0$, \emph{violated} if $\sup\mathcal I_t<0$, and \emph{inconclusive} otherwise; we call a conclusion of satisfied or violated a \emph{certificate}.
In both cases $\rho_t=\rho(\varphi,t)$, queried once at the end, $t=T$, offline and at every $t\ge\|\varphi\|$ online.
\end{problem}
For example, in the Heat and Serve mission of Section~\ref{sec:experiments_setup}, where the robot loads, heats, and serves a mug, the offline monitor waits until the mission has ended and asks once, at $t=T$, whether $\varphi_{\rm hs}$ holds at the end of the recording, i.e., whether the mug was loaded before the door was shut, heated neither too briefly nor too long, and served with the door open.
Online, it asks the same question at every step $t\ge\|\varphi_{\rm hs}\|$ of the frames recorded so far, e.g., whether the door has been opened within every 65~s so far, so that an overheating is flagged when it happens rather than after the mission.
An online certificate therefore refers to $\varphi$ at $t$ (e.g., ``not violated so far''), unlike partial-trace monitoring~\citep{deshmukh17robust}, which bounds every completion of the trace.
\vspace{-.5em}
\begin{definition}[Training, calibration sets]\label{def:datasets}
A \emph{recorded mission} $(x_{0:T},y_{0:T})$ consists of true states $x_t\in\mathcal X$ and images $y_t\in\mathcal Y$ over $t=0,\ldots,T$.
The \emph{training set} $\mathcal D_{\rm train}=\{(x^{(k)},y^{(k)})\}_{k=1}^{K}$ contains state-image pairs from recorded missions, labeled with predicate margins $\mu_i(x^{(k)})$, $i=1,\ldots,N$, to train the decoders.
For calibration, we consider two datasets, each with $n$ complete missions:
\begin{equation*}
\mathcal D_\varphi=\bigl\{\bigl(x^{(j)}_{0:T_j},y^{(j)}_{0:T_j},\mathbf r^{(j)},\hat{\mathbf r}^{(j)}\bigr)\bigr\}_{j=1}^{n},\qquad
\mathcal D_C=\bigl\{\bigl(x^{(j)}_{0:T_j},y^{(j)}_{0:T_j},\mathbf c^{(j)},\hat{\mathbf c}^{(j)}\bigr)\bigr\}_{j=1}^{n}.
\end{equation*}

\emph{Formula-level calibration:} For a fixed STL formula $\varphi$, $\mathcal D_\varphi$ contains missions of the task specified by $\varphi$, with $T_j\ge\|\varphi\|$. The vectors $\mathbf r^{(j)}$ and $\hat{\mathbf r}^{(j)}$ collect its true and estimated robustness at all queried times (Problem~\ref{prob:monitoring}), reducing to scalars $r^{(j)}$ and $\hat r^{(j)}$ offline.

\emph{Clause-level calibration:} For fixed STL clauses $C_1,\ldots,C_p$ and evaluation-time sets $\mathcal T_l(T)\subseteq\{0,\ldots,T\}$ specified before calibration, $\mathbf c^{(j)}$ and $\hat{\mathbf c}^{(j)}$ collect the true and estimated clause robustness $c_l^{(j)}(t)=\rho(C_l,t),
\hat c_l^{(j)}(t)=\hat\rho(C_l,t),$
for all $l$ and $t\in\mathcal T_l(T_j)$. These missions may differ from the monitored task, provided their clause-error scores are exchangeable with the test mission's.
\end{definition}
\vspace{-1.15em}
\begin{problem}[Calibration]\label{prob:calibration}
Given images $y_{0:t}$ of a new mission, compute interval estimates of $\rho$ calibrated at either the (a) \emph{formula-level}: from $\mathcal D_\varphi$, whose missions are exchangeable with the new one, intervals $\mathcal I_t$ that contain $\rho_t$ for every queried $t$ with probability at least $1-\alpha$; or
(b) \emph{clause-level}: from $\mathcal D_C$, whose clause errors are exchangeable with those of the new mission, clause intervals $[\hat c_l](t)$ whose I-STL robustness $[\rho](\phi,t)$ contains $\rho(\phi,t)$ with probability at least $1-\alpha$ simultaneously for all formulas $\phi$ over the clauses and all times $t$ where the required clause values lie within their calibrated time sets (Sec.~\ref{subsec:calibration}).
\end{problem}

\vspace{-1.5em}
\section{Method}\label{sec:method}
\vspace{-.25em}
Sec.~\ref{subsec:pipeline} describes predicate robustness estimation from images, Sec.~\ref{subsec:conformal} develops conformal bounds on the robustness of a given STL formula, and Sec.~\ref{subsec:calibration} extends these guarantees to all formulas composed from calibrated clauses, without requiring separate calibration for each formula.
\vspace{-0.5em}
\subsection{Latent Predicate Decoders}\label{subsec:pipeline}
We build on SLS$^2$~\citep{sls2}, which learns latent image representations by predicting future latent states from the current representation and robot actions. We retain its vision transformer (ViT) encoder and world-model training objective, which encourages the representation to capture dynamics-relevant information and prevent latent collapse. We extend this architecture with one decoder head per predicate, trained jointly with the encoder to estimate physical margins (App. \ref{app:training}).

\vspace{-3pt}
\paragraph{Encoder and decoder heads.}
To reduce occlusions, which improves predicate estimation accuracy, we combine nine camera views into a single tiled image $y_t$ (Fig.~\ref{fig:online-frames}). The encoder $E_\theta$, a ViT-S/16 initialized with self-supervised DINO weights~\citep{caron21}, maps each image to a latent representation $z_t=E_\theta(y_t)\in\mathcal Z=\mathbb R^{512}$. For each predicate $\pi_i$, we train a two-layer multilayer perceptron (MLP) $D_{\psi_i}:\mathcal Z\to\mathbb R$ to estimate its signed physical margin, $\hat\rho(\pi_i,t)=D_{\psi_i}(z_t)\approx\mu_i(x_t)=\rho(\pi_i,t)$.
Thus, the predicates share a common visual representation, while each decoder learns to estimate the physical margin associated with its predicate.

\paragraph{Physically grounded labels.}
For each predicate, we compute a signed physical margin from object poses and contact geometry, with positive values indicating satisfaction and negative values indicating violation (Appendix~\ref{app:moka}). We normalize distance-based margins by a tolerance chosen from the median margin in the final frames of successful demonstrations, and clip the resulting values to $[-1,1]$. This normalization places predicates on a common scale, where $\pm1$ indicates satisfaction or violation by at least the corresponding tolerance. The labels $\mu_i(x)$ can be obtained from simulator states, motion capture, or offline annotation.
\vspace{-.5em}
\paragraph{Co-training.}
We jointly train the encoder and all predicate decoders, allowing the latent representation to adapt to the physical quantities being estimated. Compared to freezing the encoder and training only the decoders, this greatly improves the fraction of missions for which we can certify satisfaction or violation (Section~\ref{subsec:cotrain-results}). The training objective combines three terms: (i) a squared prediction error, normalized to balance predicates with different label distributions; (ii) an invariance loss that encourages predictions to remain unchanged under variations in robot pose and unrelated objects; and (iii) the SLS$^2$ world-model loss, which encourages the latent representation to capture system dynamics (Appendix~\ref{app:training}). Adding a new predicate requires training a new decoder head and briefly fine-tuning using additional labeled data.
\vspace{-.45em}
\paragraph{Compositional robustness estimation.}
To estimate the robustness of an STL formula $\varphi$, we replace the true predicate margins in~\eqref{eq:semantics} with their decoded estimates,
$\hat\rho(\pi_i,t)=D_{\psi_i}(E_\theta(y_t))$,
and apply the remaining STL semantics unchanged. Thus, each predicate margin is estimated from the image at the corresponding time step, while temporal operators combine these estimates over the relevant time windows. Since the decoders are trained independently of any particular STL formula, we can estimate the robustness of any bounded formula over the trained predicates without retraining, using a single evaluation of the STL semantics.

\vspace{-0.65em}
\subsection{Calibration with Conformal Prediction (CP)}\label{subsec:conformal}
Since the estimated robustness may differ from its true value, it may not reliably establish satisfaction or violation. To provide probabilistic guarantees, we use split CP~\citep{angelopoulos23} to construct intervals that bound the true robustness $\rho$ by calibrating the estimation error $|\hat{\rho}-\rho|$.
Given $n$ calibration missions and a nonconformity score function fixed beforehand, let $\kappa=\lceil(n+1)(1-\alpha)\rceil$ and let $\hat q_\alpha$ be the $\kappa$-th smallest of the $n$ scores, with $\hat q_\alpha=+\infty$ if $\kappa>n$.
If the calibration missions and a new mission are exchangeable, the score $s_{\rm new}$ of the new mission satisfies
$\Pr[s_{\rm new}\le\hat q_\alpha]\ge1-\alpha$.
We treat each complete mission as one exchangeable sample, allowing prediction errors across frames within the same mission to be arbitrarily dependent.
\vspace{-.4em}
\paragraph{Formula-level calibration.}
The score of mission $j$ of $\mathcal D_\varphi$ (Problem~\ref{prob:calibration}(a)) is the decoding error of the formula, $A_j=|\hat r^{(j)}-r^{(j)}|$, which defines the radius $q_\varphi=\hat q_\alpha(A_1,\ldots,A_n)$.
A new mission with decoded robustness $\hat r$ receives the interval $\mathcal I_\varphi=[\hat r-q_\varphi,\hat r+q_\varphi]$, which solves Problem~\ref{prob:calibration}(a).
For online monitoring, the score of a calibration mission is its largest error over the queried time steps, $A_j=\max_t|\hat r^{(j)}_t-r^{(j)}_t|$, so that one radius holds at every queried time step simultaneously; the clause score~\eqref{eq:clause-score-time} takes the same maximum.
Formula-level calibration provides a potentially tighter interval, but its guarantee applies only to the specified formula $\varphi$. To simultaneously bound $m$ formulas with probability at least $1-\alpha$, a union bound requires calibrating each at level $\alpha/m$. However, when $m>(n+1)\alpha$, the resulting conformal radii become infinite, making certification impossible with only $n$ calibration missions.
\vspace{-0.7em}
\subsection{Clause-Level Intervals and I-STL Semantics}\label{subsec:calibration}
To avoid this dependence on the number of formulas, we instead calibrate the robustness errors of a shared set of clauses, allowing the resulting bounds to be reused across formulas composed from them.
We select clauses $C_1,\ldots,C_p$, each an STL formula over trained predicates, and calibrate their robustness errors using missions $\mathcal D_C$. For each clause $C_l$, we specify evaluation times $\mathcal T_l(T)\subseteq\{0,\ldots,T\}$, e.g., $\{T\}$ for offline monitoring. Let $c_l(t)=\rho(C_l,t)$ and $\hat c_l(t)=\hat\rho(C_l,t)$ denote the true and estimated clause robustness (Definition~\ref{def:datasets}). The clauses and evaluation-time rules are fixed before calibration, while formulas composed from them may be chosen afterward.
Let $s_l>0$ be a scale, e.g., the median clause error on training missions, that makes errors of clauses with different accuracies comparable (unit scales unless stated otherwise).
The score of mission $j$ is its largest scaled clause error, which defines the radius $q_C$ and the per-clause radii $\varepsilon_l$,
\begin{equation}\label{eq:clause-score-time}
  \textstyle S^{(j)}=\max_{1\le l\le p}\;\max_{t\in\mathcal T_l(T_j)}\frac{|\hat c^{(j)}_l(t)-c^{(j)}_l(t)|}{s_l},\qquad
  q_C=\hat q_\alpha\bigl(S^{(1)},\ldots,S^{(n)}\bigr),\qquad
  \varepsilon_l=q_C\,s_l.
\end{equation}
For each clause $C_l$, we construct a robustness interval $[\hat c_l](t)=[\hat c_l(t)-\varepsilon_l,\hat c_l(t)+\varepsilon_l]$ at every calibrated time $t\in\mathcal T_l(T)$. Because the calibration score is the maximum normalized error across all clauses and evaluation times within each mission, conformal prediction guarantees that these intervals simultaneously contain the true clause robustness values with probability at least $1-\alpha$.

I-STL~\citep{baird2023istl} evaluates STL over interval-valued signals by replacing each operation of~\eqref{eq:semantics} with its natural inclusion function.
Temporal operators similarly apply $\min$ or $\max$ to the interval bounds over their time windows. New STL formulas can then be constructed by treating calibrated clauses $C_1,\ldots,C_p$ as atomic predicates in~\eqref{eq:syntax}. Evaluating these formulas requires clause robustness values at time steps determined by their temporal operators (App.~\ref{app:proof}).
\vspace{-1em}
\begin{proposition}[Reusable clause intervals]\label{prop:clause-time}
Fix, before the calibration missions are observed, an error level $\alpha\in(0,1)$, clauses $C_1,\ldots,C_p$, which are STL formulas over the trained predicates, and for each clause $C_l$ a scale $s_l>0$ and a rule that assigns to every mission length $T$ a set of time steps $\mathcal T_l(T)\subseteq\{0,\ldots,T\}$.
Let the decoders be trained on data that contain neither the $n$ calibration missions nor the new mission, and let these $n+1$ missions be exchangeable.
Let $q_C$ be the radius~\eqref{eq:clause-score-time} of the calibration scores $S^{(1)},\ldots,S^{(n)}$, let the new mission, of length $T$, receive the clause intervals $[\hat c_l](t)=[\hat c_l(t)-q_Cs_l,\,\hat c_l(t)+q_Cs_l]$ for $t\in\mathcal T_l(T)$, and let $[\rho](\phi,t)$ be the I-STL robustness at $t$, on these intervals, of a formula $\phi$ in the set $\Phi$ of formulas over $C_1,\ldots,C_p$.
Let $\mathcal P(T)\subseteq\Phi\times\{0,\ldots,T\}$ denote the set of formula-time pairs $(\phi,t)$ for which all clause robustness values needed to evaluate $\phi$ at time $t$ are covered by calibration. Specifically, $(\phi,t)\in\mathcal P(T)$ if every required clause value $c_l(t')$ is evaluated at a time $t'\in\mathcal T_l(T)$.
Then, over the calibration missions and the new mission, $\Pr\bigl[\rho(\phi,t)\in[\rho](\phi,t)\ \ \forall(\phi,t)\in\mathcal P(T)\bigr]\ge1-\alpha$.
\end{proposition}
\emph{Proof sketch.}
The event $\mathcal E=\{S^{\rm new}\le q_C\}$ has probability at least $1-\alpha$ by split CP, and on $\mathcal E$ every clause value at every calibrated time step lies in its interval, whatever $\phi$ and $t$.
Every operator of~\eqref{eq:semantics} is built from negation, $\min$, and $\max$, which are monotone, so by induction over $\phi$ the true robustness stays between the propagated endpoints; Appendix~\ref{app:proof} gives the induction.\hfill$\square$

Since $\mathcal E$ does not depend on $\phi$ or $t$, both may be chosen after observing the new mission, without requiring a union bound that widens intervals significantly.
A formula is certified satisfied if the lower bound of $[\rho](\phi,t)$ is positive, violated if the upper is negative, and inconclusive when the interval $[\rho](\phi, t)$ straddles zero.
By treating each trained predicate as a clause, the method can provide robustness bounds for any bounded STL formula over these predicates, provided that all time steps needed to evaluate the formula are included in the calibrated time sets. These sets may depend on the mission length $T$ (e.g., $\mathcal T_l(T)=\{T\}$ to calibrate the final time step), provided that this dependence is fixed before calibration. This preserves exchangeability because the calibration score is computed using the same rule for every mission, including missions of different lengths.

Reusability in clause-level calibration comes at the cost of wider interval width.
For $\varphi=\bigwedge_lC_l$ with $\mathcal T_l(T)=\{T\}$ and scales $s_l=1, \forall l$, the error of every mission is at most its largest clause error, $A_j=|\min_l\hat c^{(j)}_l-\min_lc^{(j)}_l|\le\max_l|\hat c^{(j)}_l-c^{(j)}_l|=S^{(j)}$, since $\min$ is 1-Lipschitz in the maximum norm, so $q_\varphi\le q_C$.
Thus, for such conjunctions, formula-level calibration yields robustness intervals that are no wider than those obtained through clause-level calibration.

\paragraph{Binding frame.}\label{subsec:m-binding}
Since the STL robustness in~\eqref{eq:semantics} is computed using negation, minima, and maxima, its estimated value $\hat\rho(\varphi,t)$ equals the decoded robustness of some predicate $\pi_i$, possibly negated, at a particular time step $t'$. We call the predicate (possibly negated) and time step that determine the estimated formula robustness the \emph{binding literal} and \emph{binding frame}, respectively. To identify them, we trace the STL computation backward from the formula output, following the operand that determines each minimum or maximum, with a fixed rule to break ties. The resulting frame identifies the observation that determines the estimated robustness and is therefore a natural starting point for diagnosing incorrect or inconclusive certificates.

\vspace{-1em}
\section{Experiments}\label{sec:experiments_setup}\label{sec:experiments}
We test six hypotheses.
(H1) The decoded robustness is an accurate point estimate for formulas unused in training and calibration.
(H2) Formula-level calibration over complete missions yields certificates that hold on held-out missions of new scenes.
(H3) Cotraining the encoder with decoders provides statistical certificates without dramatically changing classification accuracy.
(H4) One clause-level calibration yields intervals that are reusable across the formulas composed from its clauses, without wrong certificates on Stove to Microwave, at the cost of wider intervals.
(H5) Calibrating the largest error over time yields online certificates that hold at every time step.
(H6) Physically grounded decoders certify more missions with fewer wrong certificates than learned visual STL monitors, embedding-space predicates, and predicates answered by a VLM.

\paragraph{Scenes, Missions, and Protocol}\label{subsec:setup}
We evaluate the monitor in the LIBERO simulation~\citep{libero}, where a Franka Panda arm completes two missions (Figure~\ref{fig:setup} in Appendix~\ref{app:setup}) and the monitor reads nine-view frames at 20~Hz (Figure~\ref{fig:online-frames}).
In \emph{Heat and Serve}, the robot loads a mug into a microwave, shuts the door, heats the mug for 55--65~s, and retrieves it; its formula $\varphi_{\rm hs}$~\eqref{eq:heat-serve} has four clauses: load the mug before the door shuts, heat it for at least 55~s, open the door within every 65~s, and end with the door open and the mug out.
In \emph{Stove to Microwave}, a new mission, the robot first heats a moka pot on a stove for 180--210~s and sets it down flat, and only then heats and serves the mug; its formula $\varphi_{\rm sm}$~\eqref{eq:v2} has eight clauses. Detailed definition of the two mission formulas and their predicates are in Appendix~\ref{app:formulas}.
Its four sub-missions, the pot phase, the mug phase, the timing clauses, and the handover, are conjunctions of some of these clauses (Table~\ref{tab:v2-reuse}).

In each mission, a $\pi_{0.5}$ model~\citep{pi05} performs its trained skills, and another controller completes the rest (Appendix~\ref{app:setup}).
We record 713 Heat and Serve missions and 300 Stove to Microwave missions of up to 17~min.
The Heat and Serve decoder is trained on demonstrations and rollouts of LIBERO-10 tasks and on 16 Heat and Serve missions.

All experiments use $\alpha=0.1$ and report results over 200 random splits of the missions not used for training into calibration and test sets, since the coverage of one split varies with the calibration missions it draws, with 18 and 204--206 missions for Heat and Serve and 54 and 151 for Stove to Microwave.
We report certified rate, wrong-certificate rate, and coverage, i.e., the fraction of test missions whose interval contains the true robustness.
Clause-level calibration uses unit scales unless stated otherwise.
We compare with rolling and semantic-basis monitors of \citet{hoxha26} and with Embedding Temporal Logic (ETL)~\citep{etl26}, and with a vision-language-model (VLM) monitor that passes Qwen3.5-9B~\citep{qwen35} answers per predicate through our monitor~\citep{nsvstl}, and a whole-video VLM judge.
On one H200 GPU, the encoder and all heads process one nine-view frame in \InferenceCost.

\vspace{-1em}
\section{Results}\label{sec:experiments_results}
\begin{figure}[t]
\centering
\includegraphics[width=0.9\linewidth]{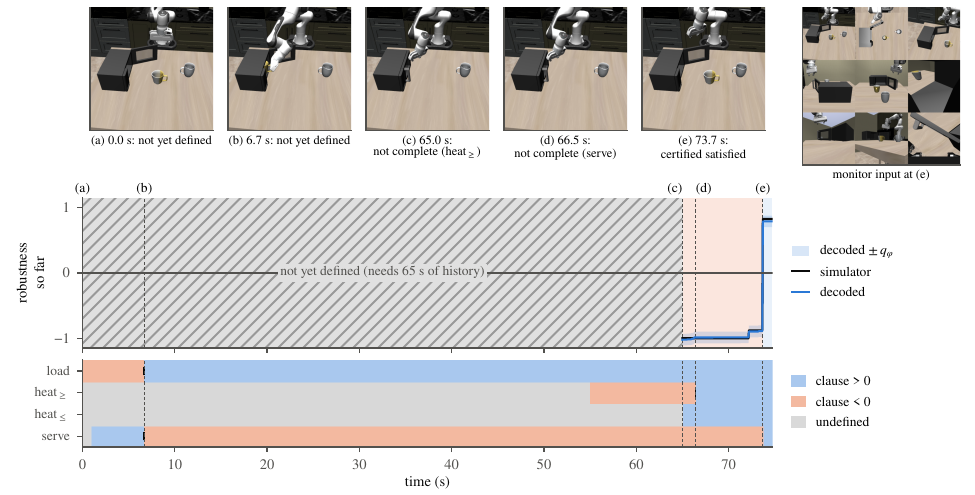}\vspace{-17pt}
\caption{Online monitoring of one Heat and Serve test mission. Top: camera view at 5 time steps with monitor's conclusion based on frames seen so far, and a nine-camera tiled image input. Middle: ground truth and decoded robustness of $\varphi_{\rm hs}$ overlaid with intervals of formula-level radius $q_\varphi$. Bottom: Sign of decoded robustness of all 4 clauses of $\varphi_{hs}$.\vspace{-15pt}}
\label{fig:online-frames}
\end{figure}

We test the six hypotheses of Section~\ref{sec:experiments_setup} in turn. Refer to Appendix~\ref{app:results} for further results.
\vspace{-.5em}
\paragraph{Point Estimates for Unseen Formulas (H1)}
We test if the decoded robustness is accurate for formulas that no training or calibration step uses (H1).
We compose the decoded predicates of Heat and Serve into 18 held-out formulas in seven classes, including nested once, three- and four-phase sequences, and response properties (Table~\ref{tab:heldout} in Appendix~\ref{app:floats}).
On 259 Heat and Serve missions not used for training, the decoded robustness has the correct sign on 4{,}609 of the 4{,}662 formula-mission pairs (98.9\%), with a median absolute error of 0.029.
On the 4{,}595 pairs with $|\rho|\ge0.1$, the sign is correct on 99.4\%, and on the 67 pairs below that value on 62.7\%; of the 53 wrong-sign pairs, 25 lie below it.
Overall, per-predicate decoders estimate the robustness of formulas chosen after training with a median error of 0.029, so such a formula can be monitored without new labels or retraining.
\vspace{-.5em}
\paragraph{Certificates on Held-Out Missions (H2)}
We test whether formula-level calibration yields correct certificates on missions of new scenes (H2).
With 18 calibration missions, formula-level calibration gives a median radius of 0.176 and certifies 96.3\% of the Heat and Serve test missions over 200 random splits, with 0.4\% wrong (Table~\ref{tab:certs} in Appendix~\ref{app:floats}).
The intervals contain the true robustness on 94.7\% of the test missions, above the nominal 90\%.
The coverage of a single split has a median of 0.964.
Formula-level calibration certifies 96.3\% of new-scene missions with 0.4\% wrong.
\vspace{-.5em}
\paragraph{Co-Training the Encoder (H3)}\label{subsec:cotrain-results}
Repeating the second training stage of the Heat and Serve decoder with the stage-1 encoder frozen keeps the per-frame predicate signs within 1.4 percentage points but widens the median formula-level radius from 0.176 to 0.754, certifying 58.7\% instead of 96.3\% of the test missions (Appendix~\ref{app:training}, Table~\ref{tab:cotrain}).
Co-training thus matters for certificates, which depend on a mission's largest error, not on per-frame sign accuracy (H3).
\vspace{-.5em}
\paragraph{Reusable Clause-Level Intervals (H4)}
We test if a single clause-level calibration provides robustness bounds that remain valid across formulas composed from the calibrated clauses, without incorrect satisfaction or violation conclusions, while quantifying the increase in interval width (H4).
On Stove to Microwave, one radius calibrated over the eight clauses (median $q_C=0.716$) certifies 51.6\% of the test missions for the whole mission and 35.3--54.3\% for its four sub-missions, with no wrong certificate (Table~\ref{tab:v2-reuse} in App.~\ref{app:floats}).
The intervals of all 255 conjunctions of the clauses contain their true robustness simultaneously on 91.0\% of the test missions, whereas a union bound over 255 formula-level radii yields infinite radii and certifies nothing.
For the whole mission, formula-level calibration has a median radius of 0.078 and certifies 96.9\% of the test missions, also without a wrong certificate; this radius is the smallest because most missions violate the whole formula at the saturated value $-1$, where decoding errors vanish (App.~\ref{app:results}).
On Heat and Serve, clause-level calibration widens the median radius from 0.176 to 0.257 and certifies 92.1\% of the test missions, with 0.3\% wrong.
Dividing each clause error by its median error on the training missions (Table~\ref{tab:phase-tubes}, clause-level with scaled errors) narrows the median width of $\varphi_{\rm hs}$ from 0.514 to 0.456 but certifies fewer missions (87.1\% against 92.1\%), because it shrinks the radii of the clauses that decide violations and enlarges that of the load clause, which decides satisfactions (App.~\ref{app:results}).
Per-predicate intervals calibrated with one shared score on Heat and Serve, applied unchanged to the load clause of Reheat, certify 98.1\% of its 182 missions with 1.0\% wrong.
No training or calibration mission contains Reheat. Although the missions are not exchangeable, transfer performs well empirically, potentially because individual clause errors remain exchangeable and independent across mission types  (App.~\ref{app:results}).
This shows one clause-level calibration covers all formulas composed from its clauses, with no wrong certificate on the longest-horizon mission.
\vspace{-.25em}
\paragraph{Online Monitoring (H5)}\label{subsec:online}
We test whether calibrating the largest error over time yields online certificates that hold at every time step (H5).
We evaluate $\varphi_{\rm hs}$ and its clauses~\eqref{eq:heat-serve} at every time step from the images recorded so far, and calibrate the score $\max_t|\hat\rho_t-\rho_t|$ (Figure~\ref{fig:online-frames}).
On Heat and Serve mission rollouts (Appendix~\ref{app:results}), the resulting intervals have a median radius of 0.216, against 0.140 for the offline score on the same missions, and contain the true robustness at every time step of 94.1\% of the test missions (Table~\ref{tab:online} in Appendix~\ref{app:floats}).
The offline radius applied at every time step covers 89.8\%, slightly below the nominal 90\%.
The intervals allow the monitor to certify 99.8\% of the mission-time-step pairs and are wrong at some time step on 0.4\% of the test missions.
On missions that complete the task, the satisfaction certificate arrives at the frame of the true completion (median delay 0~s).
The decoder therefore serves as an online monitor whose intervals hold at every time step and reaches certification at the time when the mission completes.%

\vspace{-.25em}
\paragraph{Comparison with Baselines (H6)}
We test whether physically grounded decoders certify more missions with fewer wrong certificates than existing visual monitors (H6).
The monitors of \citet{hoxha26} bound the robustness from below only, so on the Heat and Serve missions of the controller on which all baselines were run (Appendix~\ref{app:results}), their rolling monitor certifies 20.9\% and their semantic-basis monitor none of the test missions, while ours certifies 99.5\% (Table~\ref{tab:baselines}).
With their predictors retrained on the latent vectors of our frozen encoder, the semantic-basis monitor still certifies none and the rolling monitor 9.1\% of the Heat and Serve test missions.
With our two-sided calibration applied to their rolling predictor, it certifies 75.7\% of the Heat and Serve and 36.4\% of the Stove to Microwave missions with none wrong, against 99.5\% and 96.9\% for our decoder (Appendix~\ref{app:baselines}).
The one-sided calibration, not the predictor, therefore limits their monitors.
ETL~\citep{etl26} has no inconclusive outcome, so it decides, and can err on, every mission, being wrong on 29.8\% of the Heat and Serve and 32.7\% of the Stove to Microwave missions, whereas our monitor certifies 99.5\% and 96.9\% of them and is wrong on 0.4\% and 0.0\%.
With our calibration, Qwen3.5-9B predicates certify 48.0\% of the Heat and Serve test missions, with 1.5\% wrong, and 0.1\% of the Stove to Microwave test missions. 30.2\% of Qwen3.5-9B's frames deviate from the simulator by more than our formula-level conformal radius, against 11.8\% for our decoder (Figure~\ref{fig:vlm} in Appendix~\ref{app:floats}).
Under the same STL monitor and calibration, predicate estimate accuracy thus determines how many missions are certified, and it is shown here decoding physical margins gives more certified missions with fewer wrong certificates than existing visual monitors.
\begin{table}[t]
\centering\small
\caption{Baselines at $\alpha=0.1$ over 200 random splits: certified rate and wrong-certificate rate. ETL has no inconclusive outcome, so it decides, and can err on, every test mission. Heat and Serve: the missions of the one controller on which all baselines were run (Appendix~\ref{app:results}).\vspace{-10pt}}
\label{tab:baselines}
\resizebox{0.65\columnwidth}{!}{%
\begin{tabular}{l cc cc}
\hline
 & \multicolumn{2}{c}{Heat and Serve, $\varphi_{\rm hs}$ (\%)} & \multicolumn{2}{c}{Stove to Microwave (\%)}\\
\cline{2-3}\cline{4-5}
monitor & certified & wrong & certified & wrong\\
\hline
\textbf{ours, formula-level} & \textbf{99.5} & {0.4} & \textbf{96.9} & \textbf{0.0}\\
ours, clause-level & 97.4 & 0.3 & 51.6 & \textbf{0.0}\\
\hline
rolling basis from \citep{hoxha26} & 20.9 & \textbf{0.0} & 0.0 & \textbf{0.0}\\
semantic basis from \citep{hoxha26} & 0.0 & \textbf{0.0} & 0.0 & \textbf{0.0}\\
ETL \citep{etl26} & 100 & 29.8 & 100 & 32.7\\
Qwen3.5-9B predicates & 48.0 & 1.5 & 0.1 & \textbf{0.0}\\
Qwen3.5-9B whole-video judge & 35.3 & 3.8 & 40.1 & 6.2\\
\hline
\end{tabular}
}\vspace{-5pt}

\end{table}
\vspace{-.5em}
\paragraph{Summary.}
Overall, one encoder with per-predicate heads gives accurate point estimates for unseen formulas (H1), certificates for missions of new scenes (H2), intervals reusable across the formulas of a mission of up to 17~min (H4), and online certificates at every time step (H5), while existing visual monitors certify fewer missions or err more often (H6). Interestingly, we observed that representation learning done in the form of co-training the encoder with decoders changes sign accuracy very little yet yields more useful certificates (H3).
Calibrated physical predicates, composed into formulas, thus enable certified monitoring of long missions from images.

\vspace{-1.1em}
\section{Conclusion}\label{sec:conclusion}
\vspace{-0.4em}
We present a monitor that certifies STL formulas from images.
Per-predicate decoders read signed physical margins that the STL semantics compose into a point estimate for any bounded formula over the trained predicates, without retraining; on 18 held-out formulas, the sign is correct on 4{,}609 of 4{,}662 pairs. We demonstrate that co-training the encoder with decoders dramatically improves the usefulness of statistical certificates without dramatically changing classification accuracy. One conformal calibration over a set of clauses yields two-sided intervals, which, unlike the one-sided bounds of \citet{hoxha26}, also certify violation.
I-STL propagates them to every formula built from the clauses.
On the longest mission, one calibration covers all 255 conjunctions of eight clauses at once on 91.0\% of the test missions, where a union bound is vacuous.
\vspace{-.4em}
\paragraph{Limitations.}
We'd like to identify the following limitations. All experiments are conducted in simulation, and the labels need privileged state and a renderer.
Since a label is a function of poses and contact geometry only, motion capture, scripted ground truth, or offline labeling by a human or a stronger model can replace the simulator on hardware.
Because guarantee covers robustness computed from labels, label noise enters a certificate as an error that the calibration does not see; on hardware, the labels must therefore resolve margins finer than the certificates that are needed. A natural next step would be extending this work to hardware, which requires mitigating this challenge.
Clause-level intervals are 1.5 to 9 times wider than formula-level ones, showing that formula-level calibration is the right choice to obtain the tightest robustness interval estimate.

\noindent\textbf{Generative AI Use.} GenAI tools were used for coding and for writing aid.
The authors reviewed and verified all AI-assisted content and take responsibility for the final content of the paper.

\acks{This work was supported in part by the National Science Foundation under award \#2333488, and a MathWorks Microgrant.}

\bibliography{l4dc2027-sample}

\clearpage
\appendix
\section{Code Availability}
Our code is available at \url{https://github.com/gtfactslab/certify_istl}.
\section{Decoder Training}\label{app:training}
This appendix details the co-training of the encoder and the decoder heads summarized in Section~\ref{subsec:pipeline}.

\paragraph{Co-training objective.}
The encoder and all heads are trained jointly; the co-training ablation at the end of this appendix shows why.
For a batch $B$ of labeled frames $(x,y)$ of $\mathcal D_{\rm train}$, the training loss is
\begin{equation}\label{eq:train-loss}
  \mathcal L(\theta,\psi)=\frac{\sum_{(x,y)\in B}\sum_{i=1}^Nm_i(x,y)\bigl(D_{\psi_i}(E_\theta(y))-\mu_i(x)\bigr)^2/\sigma_i^2}{\sum_{(x,y)\in B}\sum_{i=1}^Nm_i(x,y)}
  +\lambda_{\rm inv}\mathcal L_{\rm inv}+\lambda_{\rm wm}\mathcal L_{\rm wm}.
\end{equation}
In words, the first term is the squared error between the decoded value $D_{\psi_i}(E_\theta(y))$ and the label $\mu_i(x)$ of every predicate $i=1,\ldots,N$, averaged over the labeled predicate-frame pairs of the batch.
The standard deviation $\sigma_i$ of $\mu_i$ over the training frames divides each error, which balances predicates with different spreads, and the mask $m_i(x,y)\in\{0,1\}$ removes frames on which $\mu_i$ is undefined or an object of $\pi_i$ is hidden in all nine views.
The world-model term $\mathcal L_{\rm wm}$ of SLS$^2$ predicts $z_{t+1}$ and $z_{t+2}$ autoregressively from $z_t$ and the robot actions with an MLP predictor and regularizes the distribution of the latent vectors against collapse; we evaluate it on windows of demonstrations, the only data with recorded actions.
The invariance term $\mathcal L_{\rm inv}$ acts on synthetic groups of five images that share the state of the objects of one predicate but differ in the arm pose and in nuisance objects (Appendix~\ref{app:moka}), and penalizes the variance of the decoded value of that predicate within a group, so that each head ignores features unrelated to its predicate.
The encoder maps the class token of the ViT (width 384) to $z_t\in\mathbb R^{512}$ through the projector of SLS$^2$, an MLP with one hidden layer of width 2048, and each head is an MLP with one hidden layer of width 256.
We use $\lambda_{\rm inv}=\lambda_{\rm wm}=1$ and AdamW with learning rates $10^{-4}$ for the encoder and $10^{-3}$ for the projector, the predictor, and the heads, and we select the checkpoint on held-out validation frames or missions.

\paragraph{Training data and new predicates.}
The Heat and Serve decoder is trained in two stages: first on 350 LIBERO-10 demonstrations (96{,}899 frames at 20~Hz) with 15 heads, one per predicate of the LIBERO-10 goals (\texttt{In}, \texttt{On}, \texttt{Close}, and \texttt{Turnon} of their objects), among them $m=\texttt{In(mug, microwave)}$ and the exact LIBERO predicate $c=\texttt{Close(microwave)}$, and then, with an added \texttt{Shut} head, on these demonstrations together with 200 $\pi_{0.5}$ rollouts of LIBERO-10 tasks and 16 Heat and Serve missions in which $\pi_{0.5}$ loads the mug and a scripted controller completes the mission.
The second stage is needed because a demonstration ends shortly after the door shuts, whereas the door stays shut for most of a heating mission.
The two stages take 8.9 and 1.3~hours on one H200 GPU.
A new predicate $\pi_j$ requires a new head $D_{\psi_j}$ and a short fine-tuning of the encoder and all heads from the previous model, rather than a new model.
To obtain training data for it, we write its predicate function $\mu_j$ as a function of the full simulator state and render synthetic scenes that vary the relevant part of the state, together with a few labeled missions (Appendix~\ref{app:moka}).
For the six moka predicates, the fine-tuning adds 35{,}000 synthetic moka images, 125{,}000 synthetic images of the LIBERO-10 scenes, and 16 moka missions, and takes 1.3~hours.
For Stove to Microwave, it further adds the 140 training missions, 92 of which contain failure states, i.e., 29 missions in which $\pi_{0.5}$ fails its skill and 63 missions in which the mug or the pot is toppled, tilted, moved off-center on the burner, or moved off the table, together with 15{,}000 synthetic images of such failure states, since a monitor must also decode the states of failed missions.

\paragraph{Co-training ablation.}
To test whether training the encoder is needed, we repeat the second stage of the Heat and Serve decoder with the same data, schedule, and checkpoint selection, but with the stage-1 encoder, i.e., the ViT and the projector, frozen, so that only the heads train; the world-model term, which acts only on the encoder, is then omitted.
We evaluate both decoders with formula-level calibration on the same 200 splits of all Heat and Serve missions (Table~\ref{tab:cotrain}).
The per-frame sign agreement of the frozen variant is lower by at most 1.4 percentage points, yet its median radius is 4.3 times larger (0.754 against 0.176), and its certified rate falls from 96.3\% to 58.7\%, while its coverage stays at 94.8\% and its wrong rate at 0.25\%.
The radius is set by the largest robustness errors of the calibration missions, which an agreement averaged over frames does not reveal.
Co-training the encoder therefore matters for the certificates rather than for the per-frame accuracy.

\section{Experimental Setup}\label{app:setup}
\begin{figure}[h]
\centering
\begin{minipage}[t]{0.32\linewidth}\centering
\includegraphics[width=\linewidth]{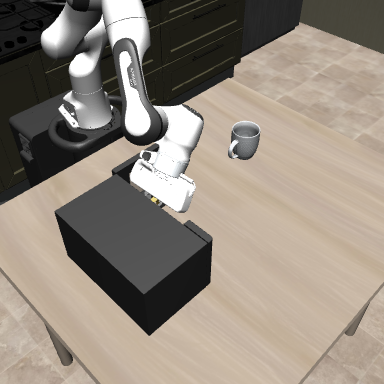}\\[1pt]
{\small (a) Heat and Serve}
\end{minipage}\hfill
\begin{minipage}[t]{0.32\linewidth}\centering
\includegraphics[width=\linewidth]{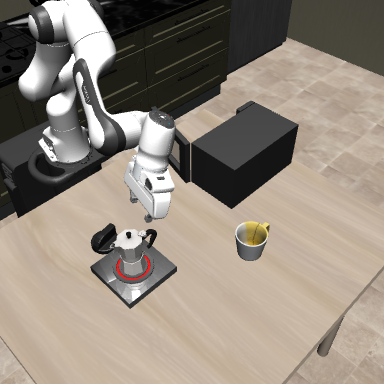}\\[1pt]
{\small (b) Stove to Microwave}
\end{minipage}\hfill
\begin{minipage}[t]{0.32\linewidth}\centering
\includegraphics[width=\linewidth]{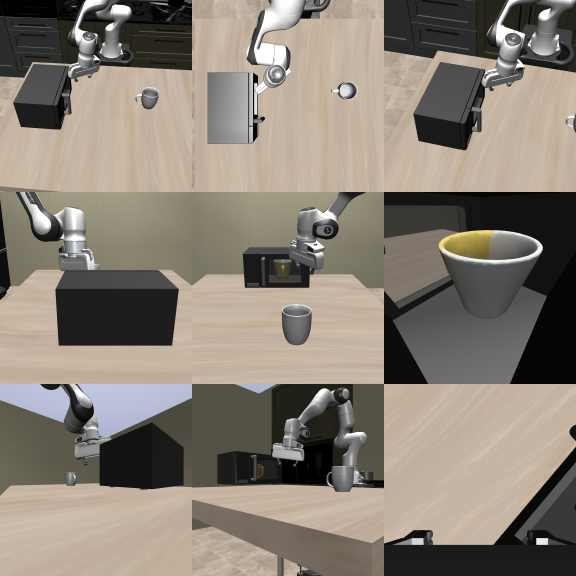}\\[1pt]
{\small (c) Monitor input}
\end{minipage}
\caption{An illustrative kitchen scene. (a) The arm loads the mug into the microwave. (b) The moka pot heats on the stove. (c) A nine-view frame sent to the encoder.}
\label{fig:setup}
\end{figure}

\paragraph{Scenes.}
In each scene, a 7-DoF Franka Panda arm with a parallel-jaw gripper stands at one edge of a kitchen table and is commanded at 20~Hz through an operational space controller (Figure~\ref{fig:setup}).
Each frame is rendered from nine cameras at $384\times384$~px, downsampled to $192\times192$~px per view, and tiled into a $576\times576$~px image, which a ViT-S/16 encoder maps to $z_t\in\mathbb R^{512}$.
The monitor reads the frames at 20~Hz, and Appendix~\ref{app:formulas} gives the mission formulas and their predicates.

\paragraph{Heat and Serve.}
The scene is the kitchen of LIBERO-10 task~9, with a microwave with a hinged door, the target mug, and a second mug as a distractor.
The robot loads the mug into the microwave, closes the door, heats the mug for 55--65~s, and then opens the door and retrieves the mug.
Its formula $\varphi_{\rm hs}=\text{load}\wedge\text{heat}_{\ge}\wedge\text{heat}_{\le}\wedge\text{serve}$~\eqref{eq:heat-serve} has four clauses: load the mug before the door is shut, keep it in the shut microwave for at least 55~s, reopen the door within 65~s, and end with the door open and the mug outside.

\paragraph{Stove to Microwave.}
This scene places a stove, a moka pot, the mug, and the microwave on one table, a combination that no LIBERO task contains.
The robot puts the moka pot on the stove for 180--210~s, sets it down flat on the table within 30~s, and only then loads, heats, and serves the mug as in Heat and Serve.
Its formula $\varphi_{\rm sm}$ is the conjunction of eight clauses~\eqref{eq:v2}: four on the pot, the clauses load, heat$_\ge$, and heat$_\le$ of Heat and Serve, and a serve clause that also requires the mug to stand upright on the table.
The four pot clauses require the pot on the burner within the first 30~s and for 180--210~s, set down flat on the table within 30~s after it leaves the burner and kept there, and the mug left on the table until the pot is set down.
Besides the whole mission, we monitor four sub-missions, each a conjunction of some of these clauses: the pot phase, the mug phase, the timing clauses, and the handover (Table~\ref{tab:v2-reuse}).

\paragraph{Recorded missions.}
In each mission, a $\pi_{0.5}$~\citep{pi05} vision-language-action model fine-tuned on LIBERO-10 performs the skills that it has been trained for, such as picking up the mug and putting the moka pot on the stove.
Afterwards, a different controller takes over, since this checkpoint does not generalize to unseen skills such as opening the microwave door and retrieving the mug from inside it.
The Heat and Serve recordings contain 713 missions: 244 from a scripted controller that sets the simulator state, 242 from a geometric planner, and 227 from model predictive control (MPC) with a state-space model; missions in which $\pi_{0.5}$ failed to load the mug, and missions whose replay diverged so that they could not be rendered, are not counted.
They last 14--202~s, and all but 16 of them last 45--160~s.
The Stove to Microwave recording contains 300 missions of 31--1018~s with a median of 429~s, 29 of which end early because $\pi_{0.5}$ fails its skill; the 63 missions with a displaced mug or pot (Appendix~\ref{app:moka}) are recorded separately and used for training only.

\paragraph{Calibration protocol.}
We split the missions at the mission level by a seeded draw into training, validation, and pool missions, and all experiments use $\alpha=0.1$.
The validation missions select the decoder checkpoint, and the pool, disjoint from both, is split at random into calibration and test sets 200 times.
We report the certified rate, the wrong-certificate rate, and the coverage, i.e., the fraction of test missions whose interval contains the true robustness, over the test missions of all 200 splits, and the median radius over the splits.
The Heat and Serve decoder is trained on LIBERO-10 demonstrations, $\pi_{0.5}$ rollouts of LIBERO-10 tasks, and 16 scripted Heat and Serve missions from the fixed initial scenes (Section~\ref{subsec:pipeline}).
Of the scripted missions, 16 train the decoder and 6 validate it, and each split draws 18 calibration and 204 test missions from the other 222, 194 of them in new scenes.
Each split of the MPC recording draws 18 calibration and 204 test missions from a pool of 222 missions, 194 of them in new scenes, and each split of the planned recording draws 18 calibration and 206 test missions from a pool of 224 missions, 196 of them in new scenes; the remaining 5 MPC and 18 planned missions of the first, fixed-scene recordings were left out when those recordings were first split and are in no pool.
For Stove to Microwave, the decoder adds the moka predicate heads (Appendix~\ref{app:moka}) and trains on 140 missions, namely 48 of the recording, its 29 failed missions, and the 63 separately recorded missions with displaced objects; 18 missions validate it, and each split draws 54 calibration and 151 test missions from the remaining 205.

\paragraph{Baselines.}
We compare with three baseline families: (i) \citet{hoxha26} with their convolutional encoder, where a rolling monitor calibrates predicted predicate values before the temporal composition, and the semantic-basis monitor predicts and calibrates the robustness of a dictionary of temporal operators over predicates; (ii) Embedding Temporal Logic (ETL)~\citep{etl26} on DINOv2~\citep{oquab2024dinov} embeddings of the nine views with its trajectory-level conformal calibration; and (iii) a VLM baseline that, following NSVS-TL~\citep{nsvstl}, prompts Qwen3.5-9B~\citep{qwen35} with one yes/no question per predicate and frame at 1~Hz, maps the probability of ``yes'' to a signed score by an isotonic regression on the training missions, and passes the scores through our STL monitor and calibration, plus a whole-video judge that uses the same model.
Appendix~\ref{app:baselines} details the baselines.

\section{Mission Formulas}\label{app:formulas}
Time bounds in the formulas below are in seconds, a formula is evaluated at the last step $T$ offline and at every step $t\ge\|\varphi\|$ online, and an operator without a time bound reaches back to the start of the mission.

\paragraph{Heat and Serve.}
The predicates $m$ and $s$ hold when the mug is inside the microwave and when the door is shut, respectively, and the mission formula is
\begin{equation}\label{eq:heat-serve}
\begin{aligned}
&\varphi_{\rm hs}=\text{load}\wedge\text{heat}_{\ge}\wedge\text{heat}_{\le}\wedge\text{serve},\quad
\text{load}=\once\bigl(m\wedge\hist_{(0,\infty)}\neg s\bigr),\\
&\text{heat}_{\ge}=\once\hist_{[0,55]}(m\wedge s),\quad
\text{heat}_{\le}=\hist\once_{[0,65]}\neg s,\quad
\text{serve}=\hist_{[0,1]}(\neg s\wedge\neg m).
\end{aligned}
\end{equation}
Here load holds once the mug has entered the microwave with the door open at every earlier step, heat$_\ge$ once some 55~s stretch has had the mug inside with the door shut, heat$_\le$ while every 65~s window contains an open-door step, and serve while the mug has been out and the door open for the last second (20 frames at 20~Hz).
The formula is defined from $t=65$~s on; at the end of a mission it requires that the door stayed open until the mug was inside, that the mug was heated for at least 55~s and the door opened within every 65~s, and that the mission ends with the door open and the mug outside.

\paragraph{Stove to Microwave.}
The predicates $h$, $f$, and $o$ hold when the pot is on the burner, flat, and on the table, and $t$ and $u$ hold when the mug is on the table and upright.
The mission formula is the conjunction of eight clauses,
\begin{equation}\label{eq:v2}
\begin{aligned}
&\varphi_{\rm sm}=\text{pot}_{\ge}\wedge\text{pot}_{\le}\wedge\text{pot\_done}\wedge\text{order}\wedge\text{load}\wedge\text{heat}_{\ge}\wedge\text{heat}_{\le}\wedge\text{serve}',\quad
\text{pot}_{\ge}=\once\hist_{[0,180)}h,\\
&\text{pot}_{\le}=\hist\once_{[0,210]}\neg h,\quad
\text{order}=\hist\bigl((f\wedge o)\vee\hist\,t\bigr),\quad
\text{serve}'=\hist_{[0,2]}(\neg s\wedge\neg m\wedge t\wedge u),\\
&\text{pot\_done}=\once_{[T-30,T]}h\wedge\hist\bigl((f\wedge o)\vee\neg\once_{[30,\infty)}\text{left}\bigr)\wedge\hist_{[0,2]}(f\wedge o),\\
&\text{left}=\neg h\wedge\once_{[\delta,\delta]}\hist_{[0,180)}h,
\end{aligned}
\end{equation}
where load, heat$_\ge$, and heat$_\le$ are those of~\eqref{eq:heat-serve} and $\delta=0.2$~s is one step at the 5~Hz rate at which this mission is monitored.
Here left holds when the pot leaves the burner after a full 180~s on it, and serve$'$, unlike serve, also requires the mug on the table and upright, over the last 2~s.
In words, the pot reaches the burner within the first 30~s of the mission, stays on it for at least 180~s and leaves it within 210~s, is set down flat on the table within 30~s after it leaves and kept there, and the mug stays on the table until the pot is set down, i.e., at every step the pot is flat on the table or the mug has been on the table throughout.
The window of $\once_{[T-30,T]}$ depends on the length $T$ of the mission.

\section{Proof of Proposition~\ref{prop:clause-time}}\label{app:proof}
We first state which clause values a formula reads, then show that the coverage event does not depend on the formula, and finally propagate the clause intervals through the formula by induction.

\paragraph{Two facts used below.}
Split conformal prediction (Section~\ref{subsec:conformal}): if the $n$ calibration missions and a new mission are exchangeable, the score $s_{\rm new}$ of the new mission and the $\kappa$-th smallest calibration score $\hat q_\alpha$, $\kappa=\lceil(n+1)(1-\alpha)\rceil$, satisfy
\begin{equation}\label{eq:conformal}
  \Pr[s_{\rm new}\le\hat q_\alpha]\ge1-\alpha.
\end{equation}
I-STL~\citep{baird2023istl} replaces every operation of~\eqref{eq:semantics} by its natural inclusion function: negation negates and swaps the endpoints, and $\min$ and $\max$ act on lower and upper endpoints separately, e.g., for $[a]=[\underline a,\overline a]$ and $[b]=[\underline b,\overline b]$,
\begin{equation}\label{eq:istl}
  [a]\wedge[b]=\bigl[\min(\underline a,\underline b),\,\min(\overline a,\overline b)\bigr],\qquad
  [a]\vee[b]=\bigl[\max(\underline a,\underline b),\,\max(\overline a,\overline b)\bigr],
\end{equation}
and a temporal operator applies $\min$ or $\max$ to the lower and upper endpoints separately over its window.

\paragraph{Query set.}
For a formula $\phi$ over the clauses $C_1,\ldots,C_p$, the \emph{query set} $\mathcal Q(\phi,t)\subseteq\{1,\ldots,p\}\times\mathbb N$ collects the pairs $(l,t')$ such that evaluating $\phi$ at time step $t$ reads the value of clause $C_l$ at time step $t'$.
It is defined recursively over the syntax~\eqref{eq:syntax} by
\begin{equation*}
  \begin{aligned}
    &\mathcal Q(C_l,t)=\{(l,t)\},\quad
    \mathcal Q(\neg\phi,t)=\mathcal Q(\phi,t),\\
    &\mathcal Q(\phi_1\wedge\phi_2,t)=\mathcal Q(\phi_1\vee\phi_2,t)=\textstyle\bigcup_k\mathcal Q(\phi_k,t),\\
    &\mathcal Q(\hist_I\phi,t)=\mathcal Q(\once_I\phi,t)=\textstyle\bigcup_{t'\in t-I}\mathcal Q(\phi,t'),\\
    &\mathcal Q(\phi_1\mathcal S_I\phi_2,t)=\textstyle\bigcup_{t'\in t-I}\bigl(\mathcal Q(\phi_2,t')\cup\bigcup_{t'<u\le t}\mathcal Q(\phi_1,u)\bigr).
  \end{aligned}
\end{equation*}
In words, a clause reads itself at $t$, negation and the Boolean connectives read what their operands read, and a temporal operator reads what its operands read at every time step of its window, including the time steps $u$ that the left operand of since covers after the window.
The pair $(\phi,t)$ is \emph{calibrated} if $\mathcal Q(\phi,t)\subseteq\mathcal K(T)$, where $\mathcal K(T)=\{(l,t'):t'\in\mathcal T_l(T)\}$ lists the calibrated clause values, i.e., if every clause value that $\phi$ reads at $t$ is calibrated.

\begin{proof}
\emph{Step 1, the coverage event.}
Let $S^{\rm new}$ be the score~\eqref{eq:clause-score-time} of the new mission.
The scores $S^{(1)},\ldots,S^{(n)},S^{\rm new}$ are one fixed function applied to exchangeable missions, where the time sets enter only through the length of each mission, so they are exchangeable.
By~\eqref{eq:conformal}, the event $\mathcal E=\{S^{\rm new}\le q_C\}$ therefore has probability at least $1-\alpha$.
Note that $\mathcal E$ is defined by the clauses, the scales, and the calibration missions alone, and not by any formula or time step.

\emph{Step 2, clause values on $\mathcal E$.}
On $\mathcal E$, the largest scaled clause error of the new mission is at most $q_C$, so $|\hat c_l(t)-c_l(t)|\le s_lS^{\rm new}\le s_lq_C=\varepsilon_l$ for all $(l,t)\in\mathcal K(T)$ at once.
That is, $c_l(t)\in[\hat c_l](t)$ for every calibrated clause value simultaneously.

\emph{Step 3, induction over the formula.}
We show that, on $\mathcal E$, $\rho(\phi,t)\in[\rho](\phi,t)$ for every calibrated $(\phi,t)$, by induction on the structure of $\phi$.
For a clause, the claim is Step~2.
For negation, $\rho(\phi,t)\in[a,b]$ implies $-\rho(\phi,t)\in[-b,-a]$, which is the interval that I-STL assigns to $\neg\phi$.
For every other operator, the robustness is $\rho(\phi,t)=f(\rho(\phi_1,t_1),\ldots,\rho(\phi_m,t_m))$, where $f$ is a composition of $\min$ and $\max$ over finitely many arguments, since a window reads only the finitely many recorded time steps, and each subformula $\phi_k$ is evaluated at a time step $t_k$ whose query set satisfies $\mathcal Q(\phi_k,t_k)\subseteq\mathcal Q(\phi,t)\subseteq\mathcal K(T)$, so that $(\phi_k,t_k)$ is calibrated.
By the induction hypothesis, $\rho(\phi_k,t_k)\in[a_k,b_k]=[\rho](\phi_k,t_k)$ for every $k$.
Since $f$ is nondecreasing in every argument, $f(a_1,\ldots,a_m)\le\rho(\phi,t)\le f(b_1,\ldots,b_m)$, and these are exactly the endpoints of $[\rho](\phi,t)$ under the natural inclusion functions~\eqref{eq:istl}.

Steps 1--3 give $\Pr[\rho(\phi,t)\in[\rho](\phi,t)]\ge\Pr[\mathcal E]\ge1-\alpha$, and since $\mathcal E$ does not depend on $(\phi,t)$, the same event covers every calibrated pair at once.
\end{proof}

\section{Additional Results}\label{app:results}
\begin{figure}[h]
\centering
\includegraphics[width=0.7\linewidth]{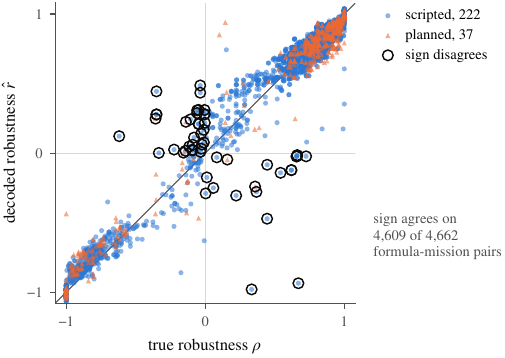}
\caption{Point estimates for the held-out formulas of Table~\ref{tab:heldout}. Decoded against true robustness on 222 scripted (194 in newly sampled scenes) and 37 planned Heat and Serve missions (those of the first planned recording, 46 missions, that had been rendered when the evaluation was run); circles mark the 53 pairs with the wrong sign.}
\label{fig:heldout}
\end{figure}

\paragraph{Held-out formulas.}
Each class has the correct sign on at least 98.2\% of its pairs (Table~\ref{tab:heldout}).
The classes cover new dwell windows, nested once, three- and four-phase sequences, response properties, invariants, bounded once with offsets and a disjunction, and formulas over a stricter door predicate $c$, which holds only when the door is fully closed.
These point estimates (Figure~\ref{fig:heldout}) require neither retraining nor calibration, and all 18 formulas reuse a single decoding pass of each mission.
Sign agreement is a lenient metric, so we also report where it fails.
Only 67 of the 4{,}662 pairs (1.4\%) have a true robustness with $|\rho|<0.1$, and 24.9\% have $|\rho|=1$, i.e., are saturated.
On the 4{,}595 pairs with $|\rho|\ge0.1$, the sign is correct on 99.4\%, and on the 67 pairs with $|\rho|<0.1$ on 62.7\%.
The 53 wrong-sign pairs have a median $|\rho|$ of 0.116, and 25 of them (47\%) have $|\rho|<0.1$.
Hence the errors concentrate where the true robustness is near zero, which an interval flags as inconclusive whenever it contains zero, and the median absolute error of 0.029 summarizes the accuracy better than the sign agreement.

\begin{figure}[h]
\centering
\includegraphics[width=\linewidth]{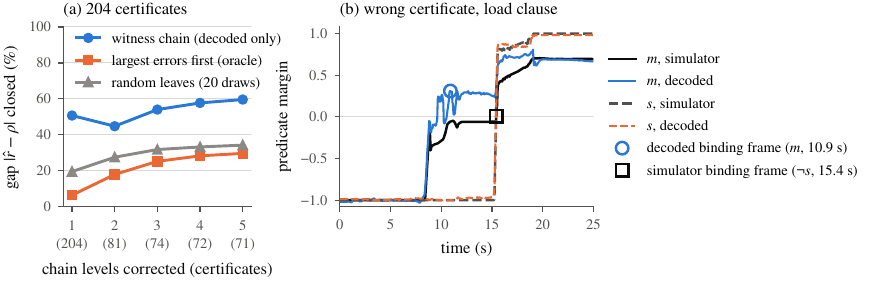}
\caption{Binding frames of $\varphi_{\rm hs}$ on the 204 Heat and Serve test missions of the scripted controller in one split. (a) Share of the gap $|\hat r-\rho|$ closed by replacing the decoded values on the first $L$ frame groups by the true values, against replacing as many literal-frame pairs with the largest decoding errors or at random (mean of 20 draws). The first group holds the binding frame and the frames of its literal whose decoded values lie within $2\cdot10^{-3}$ of $\hat r$, and group $L+1$ is found by the same procedure after the frames of groups $1,\ldots,L$ are removed, so the groups are the successive sets of frames that decide the value. In parentheses: missions with at least $L$ groups; each point is the mean over them. (b) The mission on which all wrong formula-level certificates of the scripted controller over the 200 splits fall ($\hat r=0.311$, $\rho=-0.006$): decoded and simulator margins of $m$ and $s$ in the first 25~s, with the decoded binding frame and the binding frame of the true robustness.}
\label{fig:binding}
\end{figure}

\paragraph{Binding frames.}
We test the binding frame (Section~\ref{subsec:m-binding}) on the 204 Heat and Serve test missions of the scripted controller in one split.
The binding frame of the decoded robustness lies in the same clause as the binding frame of the true robustness on 178 missions.
Replacing the decoded values of the binding literal at the binding frame and at all frames tied with it (median 40 frames) by the true values closes 51\% of the gap $|\hat r-\rho|$ on average, whereas replacing as many frames with the largest decoding errors closes 6\% (Figure~\ref{fig:binding}(a)).
Put another way, the largest decoding error is rarely the one that decides the error of the formula-level robustness, and the binding frame finds the one that does.

\paragraph{Clause-level intervals.}
Table~\ref{tab:v2-reuse} applies a single radius $q_C$, calibrated once over the eight clauses of Stove to Microwave, to the whole mission and to its four sub-missions through the I-STL semantics~\eqref{eq:istl}.
The reused intervals certify 51.6\% of the test missions, issue no wrong certificate for any of the five formulas, and cover 97.1--99.4\% of the test missions.
The intervals of all five formulas contain their true robustness simultaneously on 95.4\% of the test missions, as Proposition~\ref{prop:clause-time} requires.
The price of reuse is width.
For the whole mission, formula-level calibration has a median radius of $q_\varphi=0.078$ against $q_C=0.716$ and certifies 96.9\% of the test missions against 51.6\%, both without a wrong certificate; its coverage of 91.0\% is the expected one for $\alpha=0.1$, whereas the 99.4\% of clause-level reuse is conservative (Table~\ref{tab:certs}).
The whole-mission radius is also smaller than the formula-level radius of every sub-mission (0.092--0.235, Table~\ref{tab:v2-clauses}), because a conjunction of eight clauses is violated by most missions, 138 of the 205 in the pool, typically by a clause at the saturated value $-1$, where the decoding error vanishes: 90 pool missions have $|\rho|=1$, and on 78 of them the decoded robustness is within $10^{-3}$ of it.
The median whole-mission error is therefore 0.010, against 0.019--0.054 for the sub-missions, and the certified rate of the whole mission counts many such saturated violations.

Table~\ref{tab:phase-tubes} explains why scaling the clause errors narrows the median width of $\varphi_{\rm hs}$ (0.456 against 0.514) but certifies fewer test missions (87.1\% against 92.1\%), averaged over the three controllers.
With unit scales, every clause receives the same radius, with a median of 0.257 over the splits.
With the scales $s_l$ set to the median clause error on the training missions (load 0.046, heat$_\ge$ 0.014, heat$_\le$ 0.024, serve 0.010), the median radii become 0.625 for load, 0.188 for heat$_\ge$, 0.325 for heat$_\le$, and 0.132 for serve.
Of the 222--224 pool missions of a controller, 155--167 violate $\varphi_{\rm hs}$, and the false clause decides each of them, typically heat$_\ge$ at a robustness of $-1$, whose radius shrank, so the median width shrinks.
A satisfaction certificate, in contrast, needs every clause to clear its own radius, and the radius of the load clause more than doubled, so satisfied missions near a threshold become inconclusive: summed over the 200 splits and averaged over the controllers, the inconclusive satisfied missions rise from 1{,}143 to 3{,}617, and the load clause binds the lower endpoint in 3{,}582 of them.
Put another way, scaling moves width from the clauses that decide violations to the clause that decides satisfactions.
The radius $q_C$ is set by the clause with the largest error on the calibration missions, which on Stove to Microwave is the heat$_\ge$ clause of the mug in 145 of the 200 splits, pot\_done in 24, and order in 10; the pot phase and the handover, whose clauses are decoded more accurately, therefore receive smaller radii when calibrated over their own clauses (Table~\ref{tab:v2-clauses}) than under the shared $q_C$ (Table~\ref{tab:v2-reuse}).
A better decoder for the binding clauses, scales chosen per clause on a validation set, and more calibration missions therefore lower $q_C$, whereas a per-clause level $\alpha_l$ would reintroduce a union bound.

Intervals over time can likewise be calibrated per predicate on one mission and reused for a temporal formula on another unseen mission.
We calibrate the predicates $s$ and $m$ of Heat and Serve as two clause signals with the shared score~\eqref{eq:clause-score-time} at all frames and unit scales, on a random half, 111 missions, of the 222 pool missions of the scripted controller, so that both predicates receive one radius (median 0.354).
We apply these intervals, unchanged, through I-STL to the load clause $\once\bigl(m\wedge\hist_{(0,\infty)}\neg s\bigr)$ on all 182 missions of Reheat, a mission in the Heat and Serve scene that no training or calibration mission contains, in which the robot loads the mug, heats it twice with a door-open pause in between, and serves it after 125--141~s.
Over 200 random halves, the transferred intervals certify 98.1\% of the Reheat missions, are wrong on 1.0\%, and leave 1.9\% inconclusive; a union bound that calibrates $s$ and $m$ separately at level $\alpha/2$, finite for two predicates and 111 missions, gives radii 0.342 and 0.460 and certifies 97.1\% with 0.5\% wrong.
Since missions of different types are not exchangeable, this transfer is an empirical result and not a consequence of Proposition~\ref{prop:clause-time}.

\begin{figure}[h]
\centering
\includegraphics[width=\linewidth]{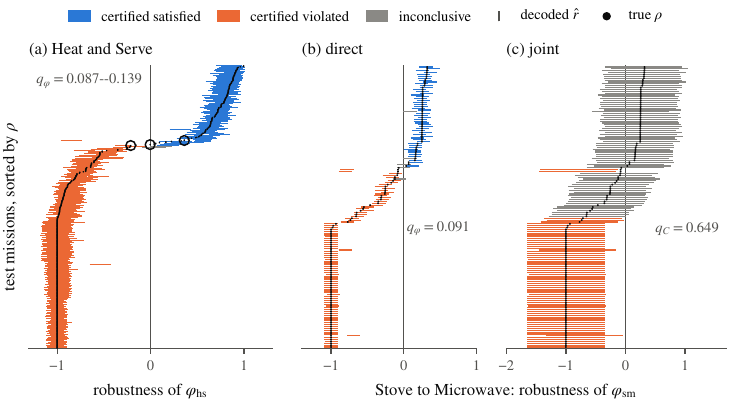}
\caption{Certificate intervals $[\hat r-q,\hat r+q]$ of the test missions of one fixed split of calibration and test sets, colored by outcome, with the decoded robustness $\hat r$ (tick) and the true robustness $\rho$ (dot); rows are sorted by $\rho$. (a) Heat and Serve, $\varphi_{\rm hs}$, formula-level calibration: the 614 test missions of the three calibrations of the split, one per controller, with radii 0.087--0.139; circles mark wrong certificates. (b)--(c) Stove to Microwave, whole mission, formula-level (direct) and clause-level (joint) calibration of the same decoded robustness, 54 calibration and 151 test missions. The other single-split figures use the same fixed split, with the 204 test missions of the scripted controller for Heat and Serve.}
\label{fig:intervals}
\end{figure}

\paragraph{Heat and Serve controllers.}
The Heat and Serve missions are completed by three controllers after $\pi_{0.5}$ has performed its trained skills: a scripted controller that sets the simulator state, a geometric planner, and MPC (Appendix~\ref{app:setup}).
The decoder is trained on 16 scripted missions, and each split calibrates and tests the missions of each controller separately; the Heat and Serve results are averages over the three controllers (Tables~\ref{tab:certs}, \ref{tab:cotrain}, and~\ref{tab:phase-tubes}), and Figure~\ref{fig:intervals}(a) shows the test missions of all three.
The comparison with the baselines, the online monitor, the binding frames, and the Reheat transfer were evaluated on the 222 pool missions of the scripted controller only, where formula-level calibration has a median radius of 0.140 and certifies 99.5\% of the test missions with 0.4\% wrong.

\paragraph{Coverage of single splits.}
The coverage reported in Table~\ref{tab:certs} is the mean over splits of the coverage of one split, which varies with the calibration missions drawn.
With $n$ calibration missions, the radius is the $\kappa$-th smallest score, and under exchangeability the coverage of one split follows a Beta$(\kappa,n+1-\kappa)$ law~\citep{angelopoulos23}.
For Heat and Serve with $n=18$, $\kappa=18$, so the law is Beta$(18,1)$ with mean 0.947, median 0.962, 5th percentile 0.846, and a probability of $0.9^{18}=0.150$ of a coverage below 0.90, i.e., 30 of 200 splits.
With formula-level calibration, single splits have a coverage between 0.686 and 1.000, with, on average over the three controllers, a 5th percentile of 0.842, a median of 0.964, and 31 of 200 splits below 0.90.
For Stove to Microwave with $n=54$, $\kappa=50$, so the law is Beta$(50,5)$ with mean 0.909, median 0.914, 5th percentile 0.838, and 0.363 probability of a coverage below 0.90, i.e., 73 of 200 splits; the formula-level splits range from 0.768 to 0.993 with 5th percentile 0.827, median 0.914, and 77 splits below 0.90.
Clause-level calibration is more conservative: its coverage per split has a minimum of 0.879 and, on average, a median of 0.985 on Heat and Serve (2.3 of 200 splits below 0.90), and a minimum of 0.967 and a median of 0.997 on Stove to Microwave (none below 0.90).
The observed spread thus matches what the theory predicts for these calibration set sizes, and a user who needs a tighter spread needs more calibration missions.

\begin{figure}[h]
\centering
\includegraphics[width=\linewidth]{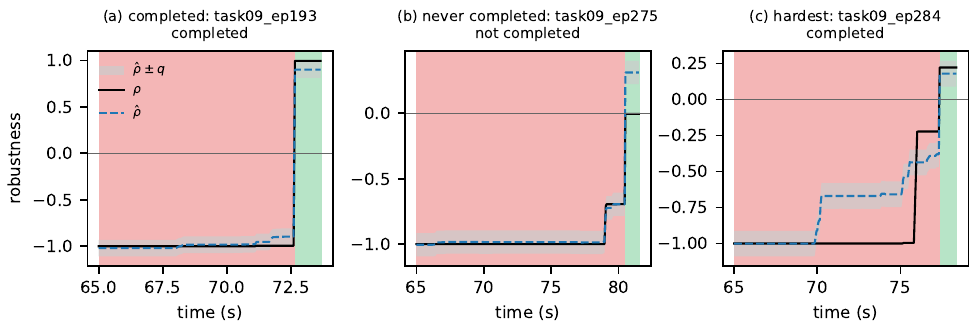}
\caption{Online intervals of $\varphi_{\rm hs}$ on three Heat and Serve test missions of one split (formula-level radius): true and decoded robustness at every time step from 65~s on, the interval, and the time ranges with a satisfaction (green) or violation (red) certificate. Left: a completed mission; middle: the mission with true robustness $-0.006$, whose decoded value ends at $0.31$, so the monitor certifies it satisfied from 80.5~s on (a wrong certificate); right: the test mission with the largest decoding error over time.}
\label{fig:online}
\end{figure}

\paragraph{Online monitoring.}
Besides $\varphi_{\rm hs}$, Table~\ref{tab:online} monitors three sub-formulas: Door and Serve, $\text{shut}_\ge\wedge\text{heat}_\le\wedge\text{serve}$ with $\text{shut}_\ge=\once\hist_{[0,55]}s$; Heat at Least, $\text{load}\wedge\text{heat}_\ge$, defined from 55~s on; and No Overheat, $\text{heat}_\le$.
A mission shorter than the look-back of a formula is never queried; for the formulas with a 65~s look-back, 23 of the 222 pool missions are, and about 183 of the 204 test missions of a split contribute queries.
We define the calibration score of a mission as its largest error over time, $\max_t|\hat\rho_t-\rho_t|$ for formula-level calibration and $\max_{l,t}|\hat c_l(t)-c_l(t)|$ over the clauses for clause-level calibration.
On the missions of the scripted controller over the 200 splits (Table~\ref{tab:online}), the formula-level intervals, with a median radius of 0.216 against 0.140 for the offline score, contain the true robustness at every time step of 94.1\% of the test missions, certify 99.8\% of the mission-time-step pairs, and are wrong at some time step on 0.4\% of the missions.
The clause-level intervals (median $q_C=0.482$) cover the true robustness on 99.4\% of the test missions, certify 85.5\% of the pairs, and are wrong on 0.1\%.
On recordings that complete the mission, the satisfaction certificate arrives at the frame of the true completion (median delay 0~s), and 98.2\% of the completed missions are certified before their end.
Figure~\ref{fig:online-frames} walks through one mission, showing at five time steps what the monitor sees and what it concludes from the frames up to that step, and Figure~\ref{fig:online} shows the online intervals on three further test missions.

\begin{figure}[h]
\centering
\includegraphics[width=\linewidth]{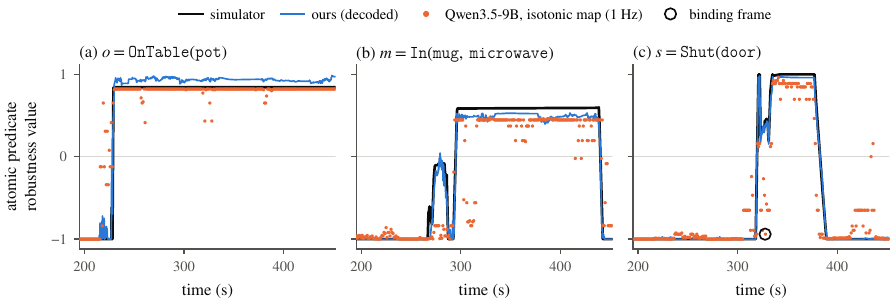}
\caption{Predicate robustness values on the satisfying Stove to Microwave test mission from 195~s to its end: simulator, our decoded margin, and the Qwen3.5-9B score under the isotonic map at 1~Hz. The circle marks the frame whose value the Qwen3.5-9B mission robustness equals. Among the test missions of this split whose true robustness is positive (54 of 151), this mission has the median mission error of Qwen3.5-9B.}
\label{fig:vlm}
\end{figure}

\paragraph{Vision-language predicates.}
On the queried Stove to Microwave test frames, 30.2\% of the values of Qwen3.5-9B deviate from the simulator margin by more than our formula-level radius, against 11.8\% for our decoder, and its sign changes lag the simulator by a median of 1.2~s, against 0.2~s (Figure~\ref{fig:vlm}).

\section{Predicates on Demand}\label{app:moka}
A new predicate enters the vocabulary through a new decoder head on the shared encoder, and the procedure of
Section~\ref{sec:method} has three steps. First, its true predicate function is written as a function of the simulator state. As for the
existing predicates, every function is a minimum of one-sided terms, each of which is a signed distance to a bound divided by a
tolerance, clipped to $[-1,1]$,
\begin{equation*}
  \mu_j(x_t)=\operatorname{clip}\Bigl(\min_k\frac{\mathrm{raw}_{j,k}(x_t)}{\mathrm{tol}_{j,k}},\,-1,\,1\Bigr).
\end{equation*}
These terms depend only on body poses and collision geometry, never on images, so that one labeling function serves both synthetic scenes
and recorded missions. Second, synthetic scenes are rendered in which the relevant part of the simulator state is varied systematically.
Third, the new head is trained on these scenes together with a few labeled missions, while the existing heads keep their training data.

\paragraph{The moka predicates.} We introduced six such predicates for a moka brew-and-pour mission in the LIBERO kitchen with a stove,
a moka pot and a mug, a combination that no LIBERO dataset contains. With $\theta$ the angle between an object's vertical axis and world
up, they are $h=\texttt{OnBurner(pot)}$ (the base center of the pot within 7.5~cm of the burner center and its lowest point within 2~cm of
the burner top), $\texttt{Tilt(pot)}$ ($\theta_{\rm pot}-45^\circ$ with a tolerance of $45^\circ$), $f=\texttt{Flat(pot)}$
($10^\circ-\theta_{\rm pot}$), $o=\texttt{OnTable(pot)}$ (every base point of the pot over the table top and its lowest point within 1~cm of
it), $\texttt{MouthOver(pot,mug)}$ (the spout tip within the 4~cm rim radius of the mug in the horizontal plane and at least 1~cm above
the rim), and $u=\texttt{Upright(mug)}$ ($30^\circ-\theta_{\rm mug}$). The Stove to Microwave mission of Section~\ref{sec:experiments_setup}
reuses $h$, $f$, $o$ and $u$ and adds one further predicate of the same form, $t=\texttt{OnTable(mug)}$.

\paragraph{Synthetic scenes.} The synthetic images are rendered in the moka scene with the camera rig of the recorded missions. They come
in groups of five images that fix the state of the pot and the mug according to one intent per predicate, such as the pot held at a sampled
tilt, the spout at a sampled offset from the rim of the mug, the pot resting on, beside or above the table, or the mug standing or toppled.
Within a group, three images share this object state but differ in the arm pose and in the placement of nuisance objects, and two are copies
of the first two in which everything except the objects of the predicate is painted out. The data comprise 35 shards, namely seven intents
in five scene variants with 200 groups each, and every image is labeled by the same predicate functions as the missions.

\paragraph{Training.} The moka model starts from the stage-1 model trained on the LIBERO-10 demonstrations, i.e., not from the Heat and Serve model, adds its own \texttt{Shut} head and the six new heads, and trains on the data of the Heat and Serve model together with the synthetic
scenes, 125{,}000 synthetic images of the LIBERO-10 scenes in groups of five, and 16 recorded moka missions. A second model is trained in the same way but without any moka
mission. The model of the Stove to Microwave mission starts from the moka model, adds the head of $\texttt{OnTable(mug)}$, and trains on
the same data together with the 48 training missions of that mission.
The decoder used for Stove to Microwave in Section~\ref{sec:experiments_results} continues training from this model on 140 training
missions together with failure states. Besides the 48 training missions, these are 29 recorded missions in which $\pi_{0.5}$ fails its skill
and 63 missions in which the mug or the pot is displaced during the mission, namely toppled, tilted or moved off-center on the burner, or
moved off the table top. It further adds 15{,}000 synthetic images in groups of five as above, which place the pot and the mug in such
failure states in frames of the training missions. No calibration or test mission is used for training.

\paragraph{Decoding accuracy.} Table~\ref{tab:moka-atoms} reports the per-frame accuracy of the six moka predicates on the test missions.
With synthetic scenes and 16 moka missions, the decoder reproduces the sign of every new predicate with a balanced accuracy of 0.97--1.00.
With synthetic scenes alone, it learns the predicates that depend on the pose of a single object (\texttt{OnBurner} 0.90, \texttt{Tilt}
0.87, \texttt{Flat} 0.92), but not the relational ones (\texttt{OnTable} 0.66, \texttt{Upright} 0.57, \texttt{MouthOver} 0.50), so a few
labeled missions are needed for the latter. With the moka heads added, the existing \texttt{Shut} head keeps a balanced accuracy of 0.999
on the microwave missions.

\begin{table}[h]
\centering\small
\caption{Moka predicates: balanced accuracy of the decoded sign and mean absolute error on the frames of the moka test missions (12{,}492 frames of 10 test missions
and 13{,}086 frames of 11 test missions at 5~Hz, since the decoder trained without moka missions splits all 50 moka missions into 39 calibration and 11 test missions), and balanced accuracy on held-out synthetic groups, for the decoder trained with synthetic scenes and 16 moka
missions and for the decoder trained without moka missions. \texttt{OnBurner} has no synthetic held-out set.}
\label{tab:moka-atoms}
\resizebox{\linewidth}{!}{\begin{tabular}{l rrr rrr}
\hline
 & \multicolumn{3}{c}{synthetic + 16 rollouts} & \multicolumn{3}{c}{synthetic only}\\
\cline{2-4}\cline{5-7}
atom & test bal.\ acc. & test MAE & synth.\ bal.\ acc. & test bal.\ acc. & test MAE & synth.\ bal.\ acc.\\
\hline
\texttt{OnBurner(pot)} & 1.00 & 0.07 & -- & 0.90 & 0.23 & --\\
\texttt{Tilt(pot)} & 0.99 & 0.13 & 0.83 & 0.87 & 0.28 & 0.70\\
\texttt{Flat(pot)} & 0.98 & 0.16 & 0.95 & 0.92 & 0.37 & 0.86\\
\texttt{OnTable(pot)} & 0.97 & 0.07 & 0.77 & 0.66 & 0.63 & 0.69\\
\texttt{MouthOver(pot,mug)} & 0.98 & 0.06 & 0.50 & 0.50 & 0.30 & 0.50\\
\texttt{Upright(mug)} & 1.00 & 0.05 & 0.81 & 0.57 & 0.19 & 0.79\\
\hline
\end{tabular}
}
\end{table}

\section{Baselines}\label{app:baselines}
All baselines use the same missions, splits, simulator labels and $\alpha=0.1$ as our monitor.

\paragraph{Rolling and semantic-basis monitors.} \citet{hoxha26} provide no public code, so we reproduced their pipeline from the paper.
Their past-time STL has no negation, so the predicate library of Heat and Serve is $\{m,\neg m,s,\neg s\}$. Each clause of
$\varphi_{\rm hs}$ is evaluated at the final frame in their implementation, and
on the original 50 missions these values agree exactly with the robustness that our monitor computes. The semantic dictionary contains the
32 depth-1 terms $\hist_I\pi$ and $\once_I\pi$ over the four literals and the four windows of the formula (1~s, 55~s, 65~s and the whole
mission), together with every clause of the Heat and Serve formulas that is not a Boolean combination of these terms, namely load, heat$_\ge$, heat$_\le$, and the clause $\text{shut}_\ge=\once\hist_{[0,55]}s$ of Door and Serve, 36 terms in total; serve is a conjunction of depth-1 terms. The rolling monitor predicts the current
predicate values, buffers them, and calibrates before the temporal composition, whereas the semantic-basis monitor predicts the robustness
of every term of the dictionary and calibrates after the composition. Both calibrate the largest scaled overestimation of the predicted
values, with scales set to the residual standard deviation on the validation missions, which yields a lower bound on the robustness of
every formula in the fragment. We use the published architecture, three convolutional layers with spatial attention per frame followed by
a temporal fusion of four frames and two-layer heads, trained from scratch on the tiled frame downsampled to $64\times64$~px. The learning
rate and the attention variant are selected by the validation loss. Because the formulas are anchored at the start of the mission, the
only valid evaluation time is the final frame, where the score is calibrated on the 18 calibration missions. On Stove to Microwave, the
dictionary contains the 196 depth-1 terms over the 14 literals and seven windows of the mission formula together with its seven clauses other than serve$'$, which is a conjunction of depth-1 terms, 203 terms in total, calibrated on the 54 calibration missions.

We also ran both monitors on our frozen encoder, with a linear layer and a ReLU per frame in place of their convolutional network and the same temporal fusion and heads (frozen variant).
The semantic-basis monitor still certifies 0.0\% of the Heat and Serve and of the Stove to Microwave test missions, and the rolling monitor certifies 9.1\% of the Heat and Serve test missions, against 20.9\% with the convolutional network.
The semantic-basis monitor calibrates the largest scaled overestimate over 36 dictionary terms (203 on Stove to Microwave), which gives a lower bound 2.5 wide on a scale of $[-1,1]$, so it cannot certify any mission.
To separate the predictor from the calibration, we apply our two-sided calibration to their rolling predictor with the convolutional network: it certifies 75.7\% of the Heat and Serve test missions with 0.0\% wrong and 36.4\% of the Stove to Microwave test missions with 0.0\% wrong, against 99.5\% and 96.9\% for our decoder.
The one-sided calibration therefore explains most of the failure of the monitors of \citet{hoxha26}.

\paragraph{Embedding Temporal Logic.} We run ETL~\citep{etl26} with the authors' code. Each predicate is the distance of the frame embedding
to a reference embedding, the mean embedding of the frames of the training missions on which the predicate holds, compared with a
per-predicate threshold. The three threshold rules are trajectory-level conformal calibration (CP$_{\rm traj}$), which calibrates the
largest score of each calibration mission, frame-level calibration (CP$_{\rm frame}$), which pools the frames on which the predicate holds,
and the threshold that maximizes the F1 score on the calibration missions ($\tau_{F1}$), which carries no guarantee. ETL decides every
mission by the sign of its formula robustness and has no inconclusive outcome, so its wrong rate equals its error rate among decided missions. All wrong rates in this paper, ours included, are over all test missions; over the certified missions only, ours are 0.43\% on Heat and Serve and 0\% on Stove to Microwave, and those of the Qwen3.5-9B predicates and judge on Heat and Serve are 3.0\% and 10.9\%. We report the trajectory-level calibration on DINOv2~\citep{oquab2024dinov} embeddings of the nine camera views.

\paragraph{Vision-language models.} Following NSVS-TL~\citep{nsvstl}, the vision-language model receives the nine camera views of a frame
as nine labeled images, each upscaled to 384~px, preceded by a description of the scene and of the camera layout, and it answers one
yes/no question per predicate (Table~\ref{tab:vlm-questions}). With $\ell_{\rm y}$ and $\ell_{\rm n}$ the next-token log-probabilities of
the answers ``yes'' and ``no'', the probability of ``yes'' is $p=e^{\ell_{\rm y}}/(e^{\ell_{\rm y}}+e^{\ell_{\rm n}})$. Frames are queried at
1~Hz, and every frame of the monitored signal takes the value of the nearest queried frame. The isotonic map is fitted per predicate on the
queried frames of the training missions. The prompts and the input layout were chosen on eight training missions before any calibration or
test frame was queried, and Qwen3.8-27B~\citep{qwen38}, a larger model of a newer generation, uses the prompts that were chosen for Qwen3.5-9B. The whole-video judge receives a time-stamped
sequence of tiled frames, one every 2~s on Heat and Serve and every 5~s on Stove to Microwave with at most 64 frames, and answers one yes/no
question per formula with its time windows stated. Its probability of ``yes'' is calibrated with conformal prediction sets: the score of a
calibration mission is one minus the probability of its true outcome, a test mission receives every outcome whose score does not exceed the
conformal quantile, and a set with a single outcome certifies that outcome.

We also ran Qwen3.8-27B with the prompts chosen for the 9B model.
On Heat and Serve, with the isotonic map, its predicates certify 32.7\% of the test missions over the 200 splits with 1.9\% wrong (45.2\% and 2.2\% with a 3~s median filter, and none with a 10~s median filter), against 48.0\% and 1.5\% for the 9B model.
On the Stove to Microwave test missions, no variant certifies any mission for the whole mission; on the 30 test missions of an earlier split with the same 54 calibration missions, the isotonic map and a 10~s median filter give a whole-mission radius of 1.164 and certify 28 of 30 missions for the pot phase, none wrong.
Asked directly for a verdict without calibration, the 27B and the 9B model both answer ``violated'' on all 204 Heat and Serve test missions of the fixed split, 62 of which satisfy $\varphi_{\rm hs}$.
The larger model therefore does not change the picture, and Table~\ref{tab:baselines} keeps the 9B model.

\begin{table}[h]
\centering\small
\caption{Questions asked to the vision-language models, one per predicate and frame. Each request consists of the scene and camera
description, the nine labeled views, one question, and ``Answer with one word: yes or no.''}
\label{tab:vlm-questions}
\begin{tabular}{l p{0.74\linewidth}}
\hline
atom & question (prompt v2) \\
\hline
OnBurner(pot) & Is the moka pot resting ON THE BURNER of the stove, i.e. its base sitting on the red ring burner, roughly centred on it (within about 7 cm), touching the burner plate (not lifted above it by more than 2 cm)? It still counts if the gripper holds it there. The close-up (3) and overhead (2) views show the burner best. \\
Flat(pot) & Is the moka pot standing UPRIGHT, its vertical axis tilted by less than 10 degrees? (Answer no if it lies on its side, leans, or is tipped in the gripper. Its location does not matter.) Look at the pot's silhouette in the front (1), side (4, 5) and close-up (3) views. \\
OnTable(pot) & Is the moka pot resting directly ON THE WOODEN TABLE TOP: its whole base over the table surface and touching it (within 1 cm), not on the stove or burner, not lifted by the gripper? The low views (7, 8) and side views (4, 5) show whether it touches the table. \\
In(mug, mw) & Is the mug INSIDE THE MICROWAVE, in its cooking cavity? The inside-microwave view (6) shows the cavity: the mug is in if it is visible there resting in the cavity. \\
Shut(mw) & Is the microwave door SHUT, i.e. closed or within about 17 degrees of fully closed? Compare the door with the microwave's front face in the front (1), overhead (2) and side (5) views; an open door sticks out from the front face at an angle. \\
Upright(mug) & Is the mug standing UPRIGHT, its axis tilted by less than 30 degrees from vertical? (Its location does not matter: table, gripper or microwave.) \\
OnTable(mug) & Is the mug resting directly ON THE WOODEN TABLE TOP: its whole base over the table surface and touching it (within 1 cm), not held up by the gripper, not inside or on top of the microwave, not on the stove? The low views (7, 8) and side views (4, 5) show whether it touches the table. \\
\hline
\end{tabular}

\end{table}

\section{Supplementary Figures and Tables}\label{app:floats}

\begin{table}[h]
\centering\small
\caption{Point estimates for held-out formulas that are never used in training or calibration, on 222 scripted (194 in newly sampled scenes) and 37 planned Heat and Serve missions (Figure~\ref{fig:heldout}). {Correct sign}: formula-mission pairs on which the decoded robustness $\hat r$ matches the sign of the true robustness $\rho$. Example time bounds are in seconds; $c$ holds when the door is fully closed. The constant formula ``heated for at least 90~s'' is omitted.}
\label{tab:heldout}
\begin{tabular}{l l c c c}
\hline
class & example & formulas & {correct sign} & median $|\hat r-\rho|$\\
\hline
dwell windows & $\once\hist_{[0,30]}(m\wedge s)$ & 3 & 763/777 & 0.028\\
nested once & $\once\bigl(s\wedge\once_{[0,10]}(m\wedge\hist\neg s)\bigr)$ & 3 & 773/777 & 0.032\\
phase sequences & load, heat $\ge40$~s, open, serve & 2 & 511/518 & 0.032\\
response & $\hist(\once_{[10,10]}s\vee\once_{[0,10]}\neg m)$ & 2 & 512/518 & 0.039\\
invariants & $\hist(m\vee\neg s)$ & 2 & 511/518 & 0.078\\
bounded once, $\vee$ & $\once_{[T-15,T-5]}m$ & 4 & 1{,}028/1{,}036 & 0.014\\
door fully closed & $\once_{[T-20,T]}(m\wedge c)$ & 2 & 511/518 & 0.024\\
\hline
all & & 18 & 4{,}609/4{,}662 & 0.029\\
\hline
\end{tabular}

\end{table}

\begin{table}[h]
\centering\small
\caption{Certificates at $\alpha=0.1$. Direct: formula-level calibration; joint: clause-level calibration with unit scales. Predicate sign agreement: per-frame agreement of the decoded and true predicate signs, $m$\,/\,$s$ over all Heat and Serve missions of a controller (including the 16 training missions), averaged over the three controllers, and the range over the seven predicates on the test frames of Stove to Microwave. Median $q$: median radius over 200 random splits into calibration and test sets ($q_\varphi$ for direct, $q_C$ for joint). Certified, wrong, and coverage: certified test missions, wrong certificates, and test missions whose interval contains the true robustness $\rho$, as percentages of the test missions of all 200 splits. Heat and Serve: averages over the three controllers, each calibrated and tested on its own missions (Appendix~\ref{app:results}).}
\label{tab:certs}
\resizebox{\linewidth}{!}{\begin{tabular}{lcl cccc}
\hline
mission & predicate sign agr.\ (\%) & scheme & median $q$ & certified (\%) & wrong (\%) & coverage (\%)\\
\hline
Heat and Serve & 99.5\,/\,99.4 & direct & 0.176 & 96.3 & 0.4 & 94.7\\
 &  & joint & 0.257 & 92.1 & 0.3 & 98.0\\
\hline
Stove to Microwave & 98.4--99.9 & direct & 0.078 & 96.9 & 0.0 & 91.0\\
 &  & joint & 0.716 & 51.6 & 0.0 & 99.4\\
\hline
\end{tabular}
}
\end{table}

\begin{table}[h]
\centering\small
\caption{Co-training ablation of the Heat and Serve decoder, $\varphi_{\rm hs}$, formula-level calibration, $\alpha=0.1$, the same 200 random splits as Table~\ref{tab:certs}. Co-trained: the decoder of all other results, whose second training stage trains the encoder and all heads; frozen: the same second stage with the stage-1 encoder frozen, so that only the heads train (Appendix~\ref{app:training}). Sign agreement, median $q_\varphi$, certified, wrong, and coverage as in Table~\ref{tab:certs}.}
\label{tab:cotrain}
\resizebox{\linewidth}{!}{\begin{tabular}{l ccccc}
\hline
encoder in stage 2 & sign agr.\ $m$\,/\,$s$ (\%) & median $q_\varphi$ & certified (\%) & wrong (\%) & coverage (\%)\\
\hline
co-trained & 99.5\,/\,99.4 & 0.176 & 96.3 & 0.4 & 94.7\\
frozen & 99.4\,/\,98.0 & 0.754 & 58.7 & 0.2 & 94.8\\
\hline
\end{tabular}
}
\end{table}

\begin{table}[h]
\centering\small
\caption{One clause-level calibration over the eight clauses of Stove to Microwave, reused for the whole mission, its four sub-missions, and all 255 conjunctions of the clauses through I-STL semantics, $\alpha=0.1$, 200 random splits into 54 calibration and 151 test missions. Pot phase: heat the pot for 180--210~s and set it down flat within 30~s. Mug phase: load, heat, and serve the mug. Timing clauses: the heating windows of the pot and the mug. Handover: set the pot down flat and keep the mug on the table until then. Coverage: test missions whose interval contains the true robustness; all five at once and all 255 conjunctions: test missions on which the intervals of all formulas contain their true robustness simultaneously; for the 255 conjunctions, certified and wrong count formula-mission pairs.}
\label{tab:v2-reuse}
\begin{tabular}{l ccc}
\hline
 & \multicolumn{3}{c}{one radius, median $q_C=0.716$ (\%)}\\
\cline{2-4}
sub-mission & certified & wrong & coverage\\
\hline
pot phase & 54.3 & 0.0 & 98.8\\
mug phase & 35.3 & 0.0 & 98.9\\
timing clauses & 40.7 & 0.0 & 97.1\\
handover & 35.9 & 0.0 & 99.1\\
\hline
\textbf{whole mission} & 51.6 & 0.0 & 99.4\\
all five at once &  &  & 95.4\\
\hline
all 255 conjunctions & 41.3 & 0.0 & 91.0\\
\hline
\end{tabular}

\end{table}

\begin{table}[h]
\centering\small
\caption{Stove to Microwave and its sub-missions at $\alpha=0.1$, 200 random splits into 54 calibration and 151 test missions. Sub-missions as in Table~\ref{tab:v2-reuse}. Median $q_\varphi$: median formula-level radius over the splits. Rates in \% for formula-level (direct) and clause-level (joint) calibration, where joint calibrates over the clauses of the respective formula only.}
\label{tab:v2-clauses}
\begin{tabular}{l cccc}
\hline
 & direct & \multicolumn{3}{c}{direct\,/\,joint (\%)}\\
\cline{3-5}
formula & median $q_\varphi$ & certified & wrong & coverage\\
\hline
pot phase & 0.235 & 95.2\,/\,90.4 & 0.0\,/\,0.0 & 91.3\,/\,94.8\\
mug phase & 0.092 & 95.1\,/\,48.4 & 0.0\,/\,0.0 & 91.5\,/\,97.4\\
timing clauses & 0.169 & 74.8\,/\,46.6 & 0.0\,/\,0.0 & 91.0\,/\,95.4\\
handover & 0.183 & 96.3\,/\,93.8 & 0.0\,/\,0.0 & 90.9\,/\,95.6\\
\hline
\textbf{whole mission} & 0.078 & 96.9\,/\,51.6 & 0.0\,/\,0.0 & 91.0\,/\,99.4\\
\hline
\end{tabular}

\end{table}

\begin{table}[h]
\centering\small
\caption{Calibration schemes on the same Heat and Serve predictions, $\alpha=0.1$, 200 random splits into 18 calibration and 204--206 test missions, averaged over the three controllers. Joint, scaled: clause-level calibration with each clause error divided by its median on the training missions. Union bound: one formula-level radius per clause at level $\alpha/4$, which is infinite with 18 calibration missions. Median width: median over the splits of the median interval width of $\varphi_{\rm hs}$ over the test missions. Certified, wrong, and coverage: percentages of the test missions of all splits. The coverage of the sub-formulas load$\,\wedge\,$heat$_\ge$ and heat$_\le$ uses the radii of the row as calibrated for $\varphi_{\rm hs}$, except in the direct row, where each sub-formula has its own formula-level radius.}
\label{tab:phase-tubes}
\resizebox{0.9\linewidth}{!}{\begin{tabular}{l ccc ccc}
\hline
 & \multicolumn{3}{c}{$\varphi_{\rm hs}$} & \multicolumn{3}{c}{coverage (\%)}\\
\cline{2-4}\cline{5-7}
scheme & median width & certified (\%) & wrong (\%) & $\varphi_{\rm hs}$ & load$\,\wedge\,$heat$_\ge$ & heat$_\le$\\
\hline
direct (formula-level) & 0.352 & 96.3 & 0.4 & 94.7 & 94.4 & 94.7\\
joint (clause-level) & 0.514 & 92.1 & 0.3 & 98.0 & 96.7 & 99.5\\
joint, scaled & 0.456 & 87.1 & 0.2 & 96.5 & 97.8 & 99.7\\
union bound over clauses & $\infty$ & 0.0 & 0.0 & 100.0 & 100.0 & 100.0\\
\hline
\end{tabular}
}
\end{table}

\begin{table}[h]
\centering\small
\caption{Online monitoring of Heat and Serve with the formulas~\eqref{eq:heat-serve}, evaluated at every time step from 65~s on, $\alpha=0.1$, 200 random splits of the missions of the scripted controller into 18 calibration and 204 test missions. Formula-level: one radius per formula from the score $\max_t|\hat\rho_t-\rho_t|$; clause-level: one radius $q_C$ from the largest clause error over time, shared by all formulas. Coverage: test missions whose intervals contain the true robustness at every time step; certified: mission-time-step pairs with a conclusion; wrong: test missions with a conclusion of the wrong sign at some time step; completed certified: missions whose true robustness becomes positive and that are certified satisfied before their end; delay: median time from the true completion to the satisfaction certificate. The last block applies the radius of the offline, end-of-mission score at every time step, which that score does not calibrate.}
\label{tab:online}
\resizebox{\linewidth}{!}{\begin{tabular}{l ccccc c}
\hline
 & $q$ & coverage & verdict & wrong & completion & delay\\
 & & (\%) & pairs (\%) & any $t$ (\%) & certified (\%) & (s)\\
\hline
\multicolumn{7}{l}{\emph{direct (formula level)}}\\
Heat and Serve & 0.216 & 94.1 & 99.8 & 0.4 & 98.2 & 0.0\\
Door and Serve & 0.147 & 94.1 & 99.6 & 0.0 & 99.2 & 0.0\\
Heat at Least & 0.286 & 94.6 & 90.1 & 2.0 & 81.5 & 0.1\\
No Overheat & 0.188 & 94.3 & 97.2 & 0.9 & 99.8 & 0.0\\
\hline
\multicolumn{7}{l}{\emph{joint (clause level, one $q_C$)}}\\
Heat and Serve & 0.482 & 99.4 & 85.5 & 0.1 & 69.0 & 0.0\\
Door and Serve & 0.482 & 99.6 & 85.9 & 0.0 & 83.9 & 0.0\\
Heat at Least & 0.482 & 97.8 & 82.5 & 1.3 & 69.5 & 0.1\\
No Overheat & 0.482 & 99.0 & 91.1 & 0.6 & 94.8 & 0.0\\
\hline
\multicolumn{7}{l}{\emph{reference: offline score $|\hat\rho_T-\rho_T|$ applied at every $t$ (not calibrated over $t$)}}\\
Heat and Serve & 0.140 & 89.8 & 99.9 & 0.4 & 99.1 & 0.0\\
Door and Serve & 0.117 & 86.5 & 99.6 & 0.0 & 99.8 & 0.0\\
Heat at Least & 0.201 & 82.6 & 97.5 & 2.8 & 95.4 & 0.1\\
No Overheat & 0.038 & 72.4 & 100.0 & 3.6 & 100.0 & 0.0\\
\hline
\end{tabular}
}
\end{table}

\FloatBarrier

\end{document}